\pdfoutput=1

\documentclass[11pt]{article}

\usepackage[final]{coling}

\usepackage{times}
\usepackage{latexsym}

\usepackage[T1]{fontenc}

\usepackage[utf8]{inputenc}

\usepackage{microtype}

\usepackage{inconsolata}
\usepackage{adjustbox}
\usepackage{amsmath}
\usepackage{amssymb}
\usepackage{dsfont}
\usepackage{subfigure}
\usepackage{graphicx}
\usepackage{CJKutf8}
\usepackage{caption}
\usepackage{booktabs}
\usepackage{siunitx}
\usepackage{wrapfig}
\usepackage{rotating}
\usepackage{float}
\usepackage{bm}
\usepackage{hyperref}
\usepackage{multirow}
\usepackage{tabu}
\usepackage{diagbox}
\usepackage{algorithm}
\usepackage{algpseudocode}

\def\h{{\boldsymbol h}}

\usepackage{graphicx}

\title{Understanding the RoPE Extensions of Long-Context LLMs: \\
An Attention Perspective}

\author{
    Meizhi Zhong\textsuperscript{\rm 1}\thanks{Work during Xiaohongshu internship.},\space\space
    Chen Zhang\textsuperscript{\rm 2},\space\space
    Yikun Lei\textsuperscript{\rm 2}, \space\space
    Xikai Liu\textsuperscript{\rm 2},\space\space 
    Yan Gao\textsuperscript{\rm 2},\space\space 
    Yao Hu\textsuperscript{\rm 2},\space\space \\
    \textbf{Kehai Chen}\textsuperscript{\rm 1}\thanks{Corresponding authors},\space\space 
    \textbf{Min Zhang}\textsuperscript{\rm 1}\space\space \\
    \textsuperscript{\rm 1}Institute of Computing and Intelligence, Harbin Institute of Technology, Shenzhen, China \\
    \textsuperscript{\rm 2} Xiaohongshu Inc.\\
    {\tt meizhi.zhong.1999@gmail.com},\space\space
    {\tt chenzhang9702@outlook.com},\space\space \\
    {\tt \{chenkehai,zhangmin2021\}@hit.edu.cn},\space\space \\
    {\tt \{zhizhu,xikai,yadun,xiahou\}@xiaohongshu.com}
}

\begin{document}
\maketitle
\begin{abstract}

Enabling LLMs to handle lengthy context is currently a research hotspot. 
Most LLMs are built upon rotary position embedding (RoPE), a popular position encoding method. 
Therefore, a prominent path is to extrapolate the RoPE trained on comparably short texts to far longer texts. 
A heavy bunch of efforts have been dedicated to boosting the extrapolation via extending the formulations of the RoPE, however, few of them have attempted to showcase their inner workings comprehensively. 
In this paper, we are driven to offer a straightforward yet in-depth understanding of RoPE extensions from an attention perspective and on two benchmarking tasks. 
A broad array of experiments reveals several valuable findings: 1) Maintaining attention patterns to those at the pretrained length improves extrapolation; 2) Large attention uncertainty leads to retrieval errors; 3) Using longer continual pretraining lengths for RoPE extensions could reduce attention uncertainty and significantly enhance extrapolation.

\end{abstract}

\section{Introduction}

Large language models (LLMs) \citep{radford2018improving,touvron2023llama,zhang2023towards,li2024tf,zhang2024modification,zhang2024dynamic} have accommodated a wide range of natural language processing applications, such as code completion \citep{rozière2023code} and question answering \citep{kamalloo2023evaluating,jiang2021can,su2019generalizing}. 
However, a notable challenge limiting further customization is possibly the inability of LLMs to utilize context beyond the pretrained length \citep{minaee2024large,chen2023extending} due to the inherent flaw of rotary position embedding (RoPE) being used.
Fortunately, RoPE extensions emerge as key ingredients to enabling LLMs to leverage extended context that exceeds pretrained scope \citep{chen2023extending,peng2023yarn,liu2023scaling,han2023lm,rozière2023code}. 
These RoPE extensions focus on improving performance on long texts, yet frustratingly, only a few of them \cite{liu2023scaling,han2023lm,men2024base} have explored the underlying mechanisms in depth.

Thus, we systematically analyze common RoPE extensions more straightforwardly, from the perspective of attention \citep{vaswani2017attention}. We include three widely-used RoPE extensions, i.e., position interpolation~\citep{chen2023extending}, YaRN~\cite{peng2023yarn}, and NTK-Aware interpolation~\citep{rozière2023code}. To our best knowledge, there is simply no research in {\em understanding RoPE extensions for long-context models thoroughly from an attention perspective}.

As a start, we strive to primarily study these methods on a long-context perplexity test (PPL) and empirically compare their corresponding attention patterns.
We found that finetuning LLMs with these RoPE-extension methods which match the original pretraining length improves extrapolation performance. Particularly with the NTK-Aware interpolation method, one can extrapolate up to 32$\times$ beyond the pretrained length.
To unleash the reasons behind the successes of these methods, we collect the attention scores respectively distributed in 2K and 8K lengths during inference. 
The results demonstrate that these methods maintain attention patterns consistent with those observed at the pretrained length.
In contrast, the attention patterns of the RoPE are substantially deviated. 

Afterward, following literature~\citep{fu2024data}, we examine these RoPE extensions on a more challenging long-context test called Needle-in-a-Haystack (Needle)~\citep{needleinhaystack}.
We find that the RoPE extensions could pass more tests than the RoPE does. 
Nonetheless, as the context length increased, the RoPE extensions could hardly locate the needles. 
We associate the observation with attention uncertainty.
We uncover that large uncertainty leads to retrieval errors: the positions that incur large attention uncertainty are exactly where the incorrect answers are borrowed from. 

We further hypothesize that this large attention uncertainty stems from a mismatch between the context lengths in training and inference.
Inspired by the conjecture, a natural way to ease the mismatch is to directly train on longer texts. 
Experimental results exhibit that, with the same amount of training tokens consumed, using examples with longer contexts largely alleviates uncertainty. Thereby, the ability to digest long texts is promoted.

Our key contributions can be summarized as follows:
\begin{itemize}
    \item We study various RoPE extensions for length extrapolation in perplexity testing and find that the effectiveness could be yielded from maintaining the original attention patterns.

    \item We analyze these methods using advanced Needle testing and observe that they may fail to extrapolate to regions where large attention uncertainty persists.

    \item We hypothesize that large attention uncertainty stems from a context length mismatch between training and inference. It is possible to reduce this large uncertainty by minimizing the mismatch through continual training with lengths closer to those in inference.

\end{itemize}

\section{Backgroud}

\subsection{Target LLMs}

We consider LLaMa series at different sizes to conduct experiments, including MiniMA-2-3B \citep{zhang2023towards}, LLaMa-2-7B, and LLaMa-2-13B \citep{touvron2023llama}.
All these mentioned LLMs consistently use rotary position embeddings to take position information into consideration.
Owing to space limitation, we only present the experimental results for LLaMa-2-7B, and the results for MiniMA-2-3B and LLaMa-2-13B, share similar trends with those for LLaMa-2-7B, as shown in Appendix \ref{minima_2_3b} and \ref{llama_2_13b}.

\subsection{RoPE and Its Extensions}

\textbf{Rotary Position Embedding (RoPE).}

Before diving into RoPE extensions, we first briefly describe RoPE itself. 
The use of RoPE~\citep{su2021roformer} has become pervasive in contemporary LLMs~\citep{touvron2023llama,bai2023qwen,bi2024deepseek}. RoPE encodes the position information of tokens with a rotation tensor that naturally incorporates explicit relative position dependency. To illustrate, given a hidden vector $\h=[h_0,h_1,...,h_{d-1}]$, where $d$ is the hidden dimension, and a position index $m$, RoPE operates as follows:
\begin{equation}
\resizebox{0.89\linewidth}{!}{$
	f(\h,m) = 
	\begin{pmatrix}
		h_0\\
		h_1\\
		h_2\\
		h_3\\
		\vdots\\
		h_{d-2}\\
		h_{d-1}
	\end{pmatrix}
	\otimes
	\begin{pmatrix}
		\cos{m\theta_0} \\
		\cos{m\theta_0} \\
		\cos{m\theta_1} \\
		\cos{m\theta_1} \\
		\vdots \\
		\cos{m\theta_{d/2-1}} \\
		\cos{m\theta_{d/2-1}} 
	\end{pmatrix}
	+
	\begin{pmatrix}
		-h_1\\
		h_0\\
		-h_3\\
		h_2\\
		\vdots\\
		-h_{d-1}\\
		h_{d-2}
	\end{pmatrix}
	\otimes
	\begin{pmatrix}
		\sin{m\theta_0}\\
		\sin{m\theta_0}\\
		\sin{m\theta_1}\\
		\sin{m\theta_1}\\
		\vdots\\
		\sin{m\theta_{d/2-1}}\\
		\sin{m\theta_{d/2-1}}
	\end{pmatrix}$}
\label{eq_rope}
\end{equation} 
where $\theta_j=b^{-2j/d},j\in\{0,1,...,d/2-1\}$, and $b$ represents the base frequency for RoPE.

\textbf{Position Interpolation (PI).}
As described in \citet{Chen2023ExtendingCW} and \citet{kaiokendev}, PI involves proportionally downscaling the position index $m$ to $m/\alpha$ in Equation \ref{eq_rope}.

\textbf{NTK-Aware Interpolation (NTK).}
NTK~\citep{rozière2023code} assumes that interpolating all dimensions equally, as done by PI, may result in the loss of high-frequency information. Therefore, NTK introduces a nonlinear interpolation strategy by adjusting the base frequency $b$.

\textbf{Yet another RoPE extensioN (YaRN).}
Unlike PI and NTK, which treat each dimension of RoPE uniformly, YaRN~\citep{peng2023yarn} employs a ramp function to combine PI and NTK at varying proportions across different dimensions. Additionally, it introduces a temperature factor to mitigate the distribution shift of the attention caused by long inputs.

Following the default settings of the original papers \citep{chen2023extending, peng2023yarn, liu2023scaling}, we adjust $\alpha$ from 1 to 16 in $m/\alpha$ for \textbf{PI} and \textbf{YaRN}, while adjusting $b$ from 10,000 to 1,000,000 for \textbf{NTK} in our experiments.

\subsection{Long-Context Evaluations}
Following existing works~\citep{chen2023extending,peng2023yarn,fu2024data}, we use the perplexity test (dubbed PPL) as the primary evaluation and the Needle-in-a-Haystack test as a more challenging evaluation.
The perplexity is a primary measure that reflects a model's ability to handle long texts. 
The Needle-in-a-Haystack test (dubbed Needle) \citep{needleinhaystack} requires LLMs to accurately recall a specific sentence (the Needle) embedded at an arbitrary location within a long document (the haystack). 
We obtain the perplexity on the Proof-pile \citep{proofpile} dataset. We follow the standard described in \citet{fu2024data} for the Needle-in-a-Haystack accuracy.

\begin{figure*}[t]
  \centering
  \subfigure[LLaMa-2-7B]{\includegraphics[width=0.3\textwidth]{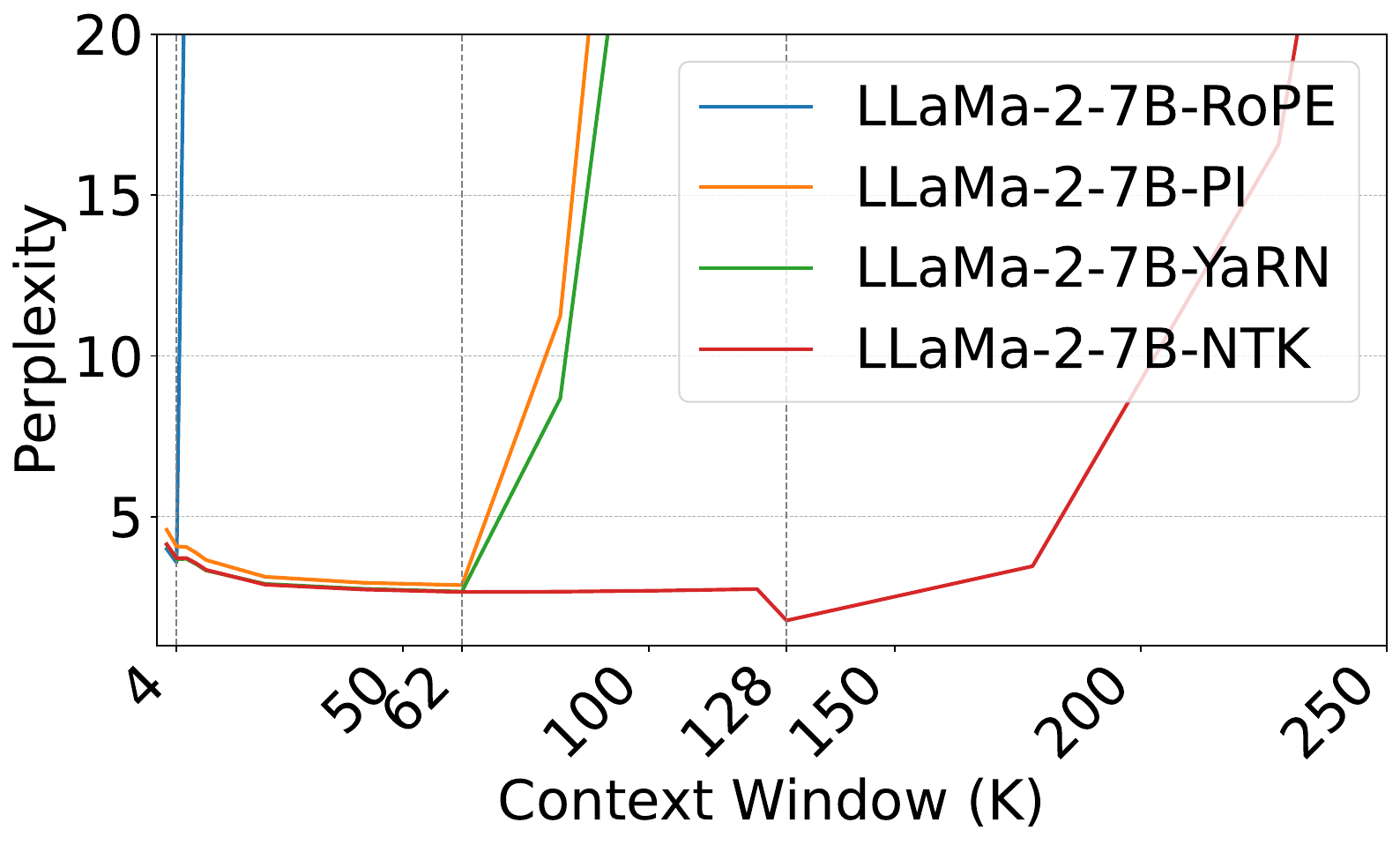}\label{fig_val_ppl_llama2}}
  \subfigure[MiniMA-2-3B]{\includegraphics[width=0.3\textwidth]{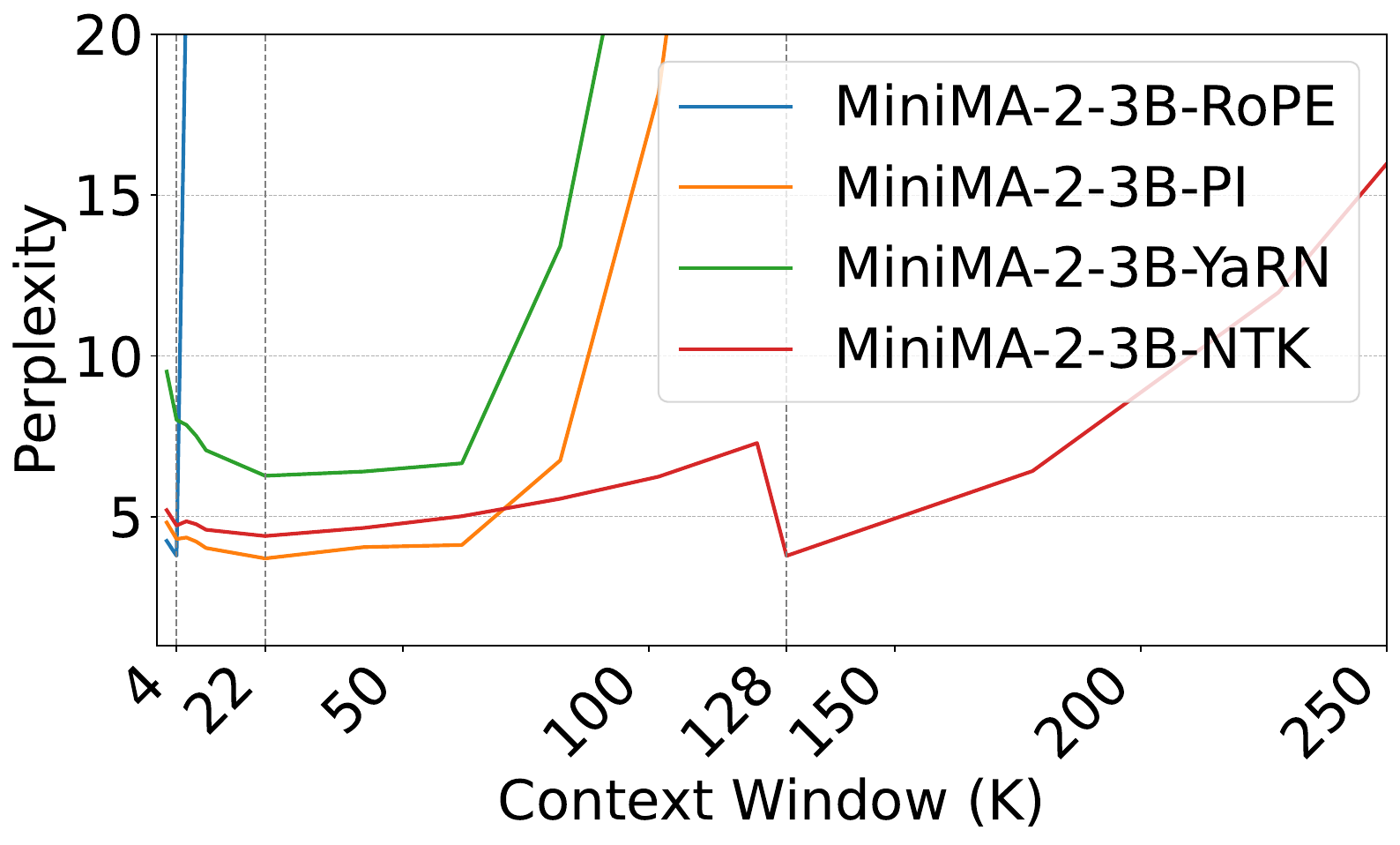}\label{fig_val_ppl_minima}}
  \subfigure[LLaMa-2-13B]{\includegraphics[width=0.3\textwidth]{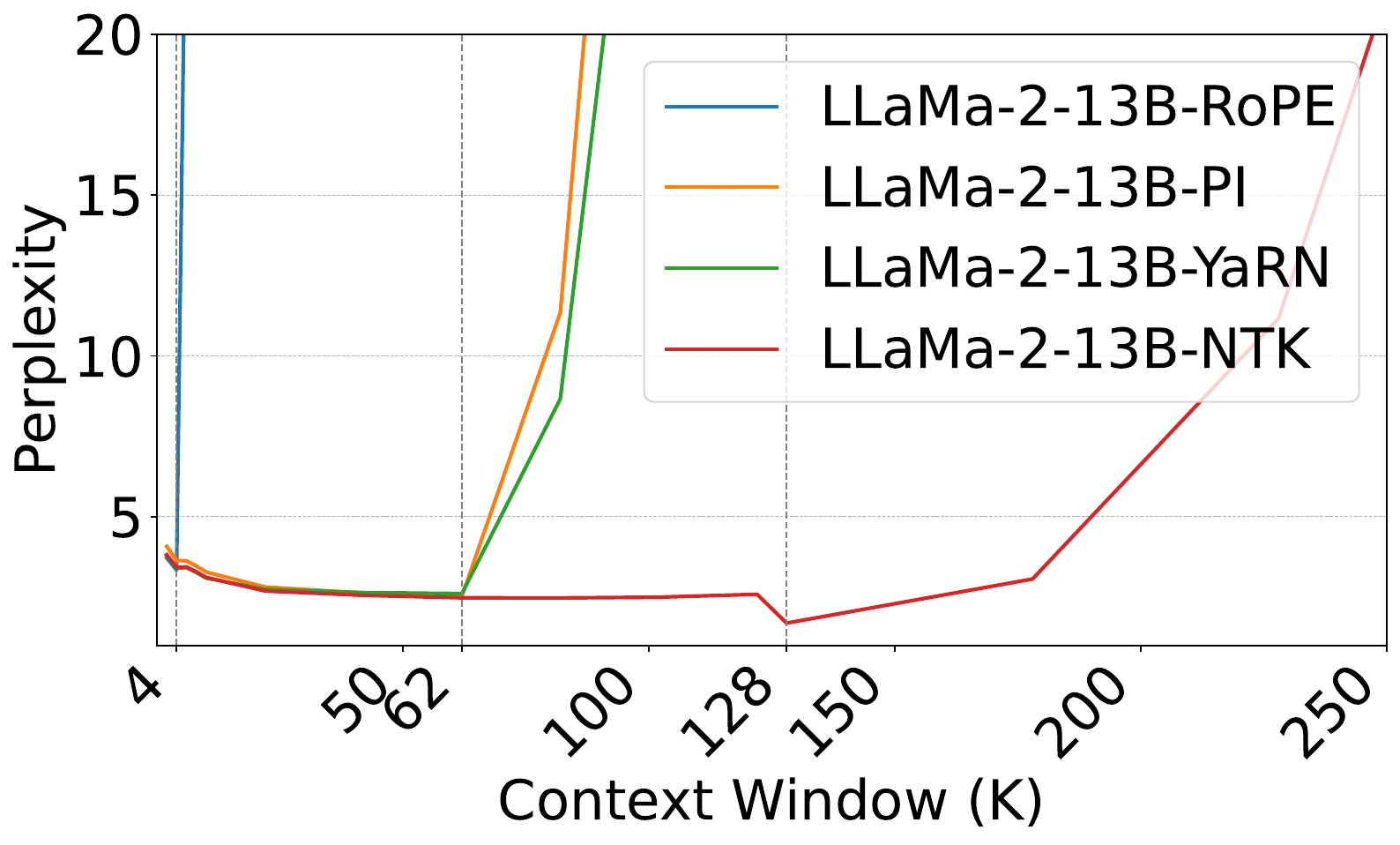}\label{fig_val_ppl_llama2_13b}}
  \vspace{-3mm}
  \caption{Perplexity on Proof-pile(Lower is better).}
  \label{fig_val_ppl}
\end{figure*}

\section{RoPE Extensions on PPL}
\label{rope_on_ppl}
\begin{figure*}[!htbp]
    \centering
    \subfigure[RoPE on 2K sequences.]{\includegraphics[width=0.315\textwidth]{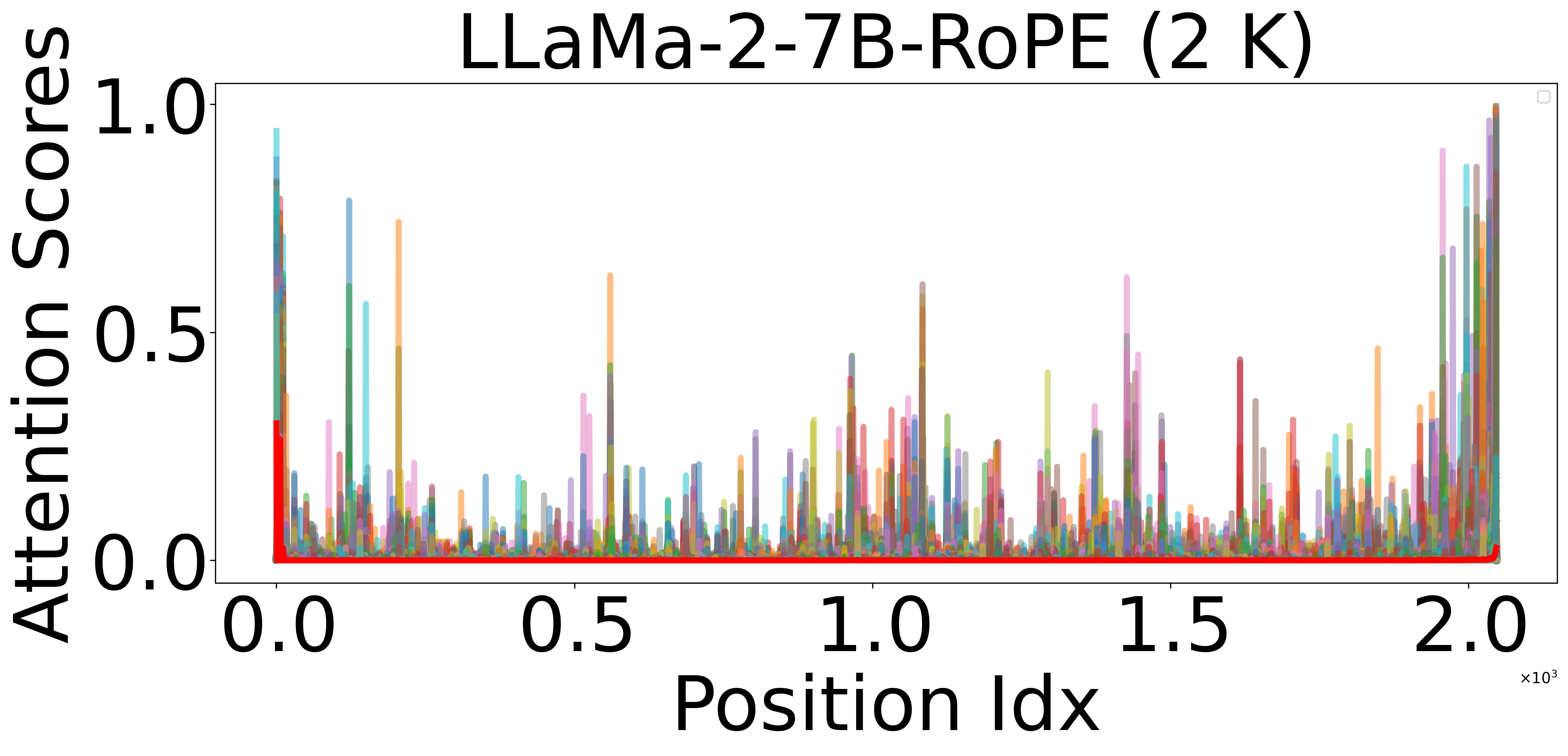}\label{fig_7b_2k_attn}}
    \subfigure[RoPE on 8K sequences.]{\includegraphics[width=0.3\textwidth]{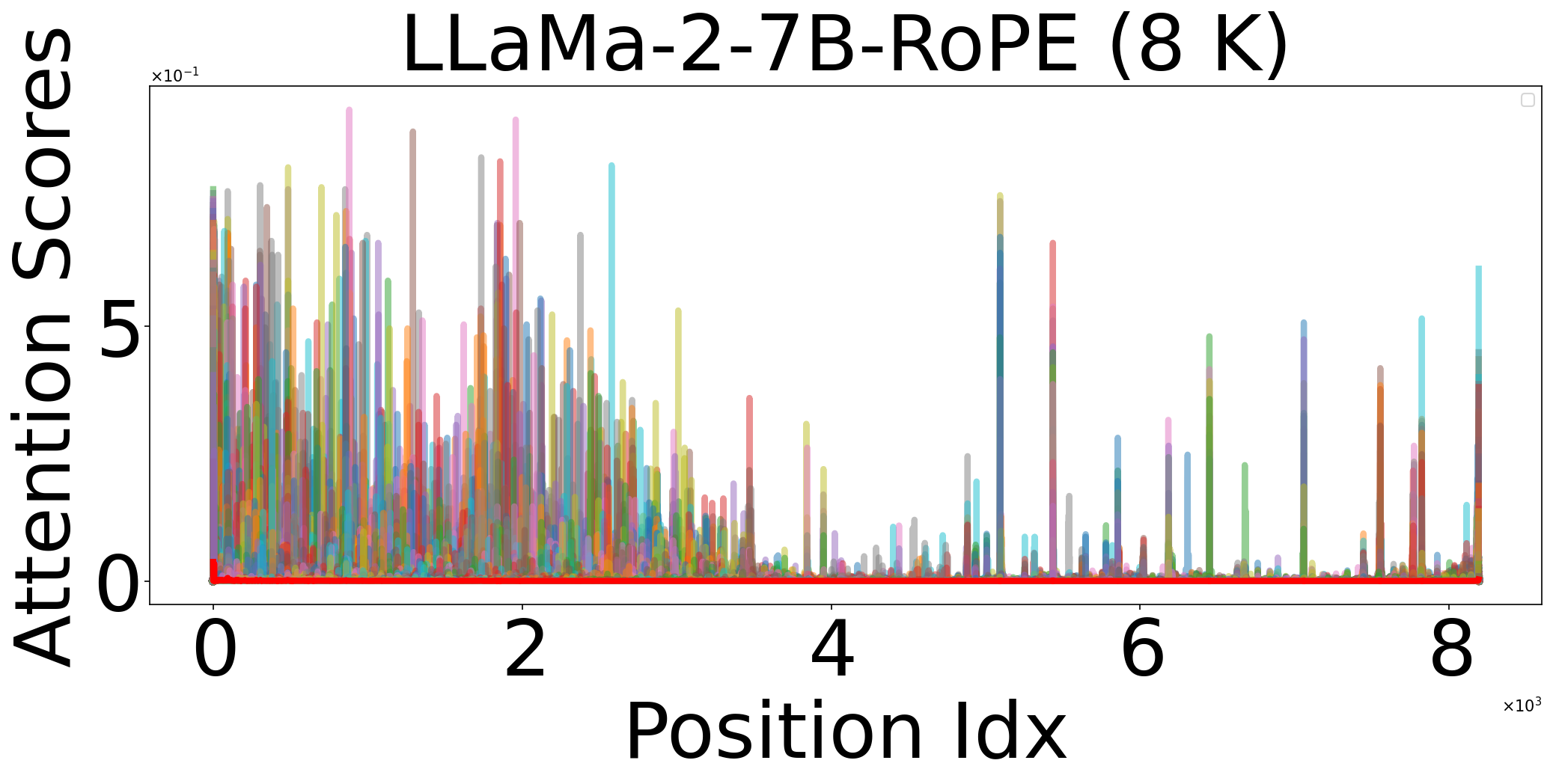}\label{fig_7b_8k_attn}}
    \subfigure[PI on 8K sequences.]{\includegraphics[width=0.315\textwidth]{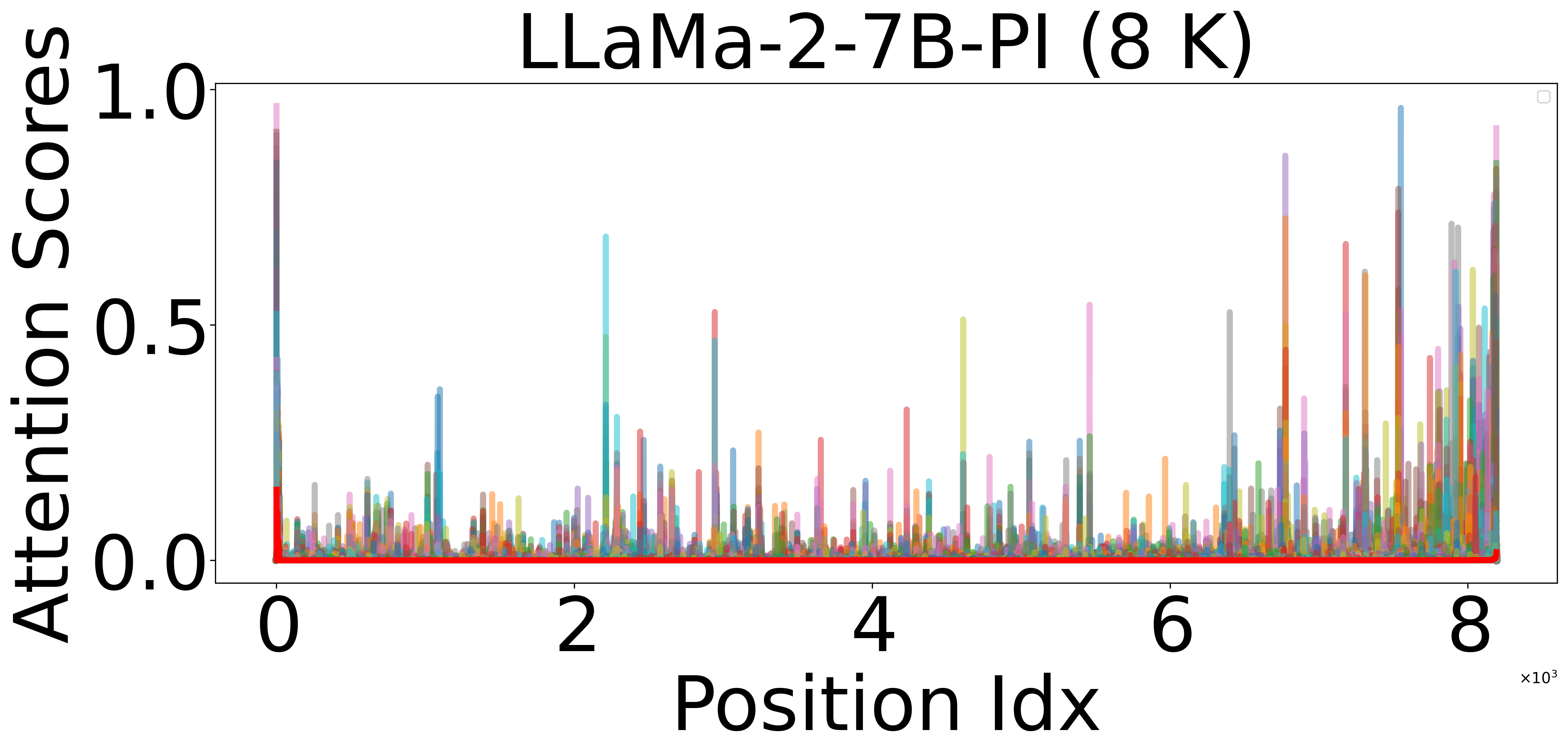}\label{fig_pi_7b_8k_attn}
    }
    
    \subfigure[YaRN on 8K sequences.]{\includegraphics[width=0.32\textwidth]{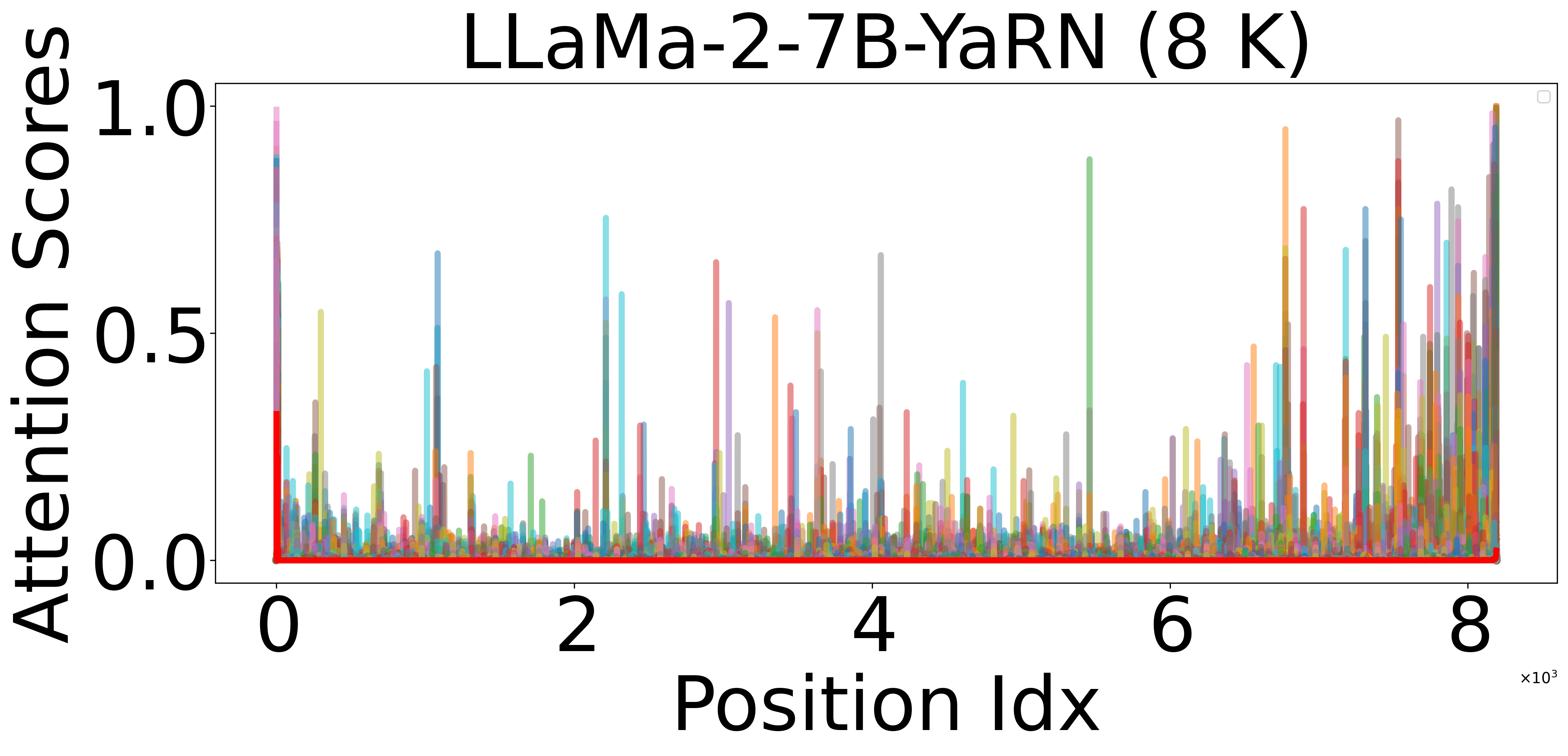}\label{fig_yarn_7b_8k_attn}}
    \subfigure[NTK on 8K sequences.]{\includegraphics[width=0.32\textwidth]{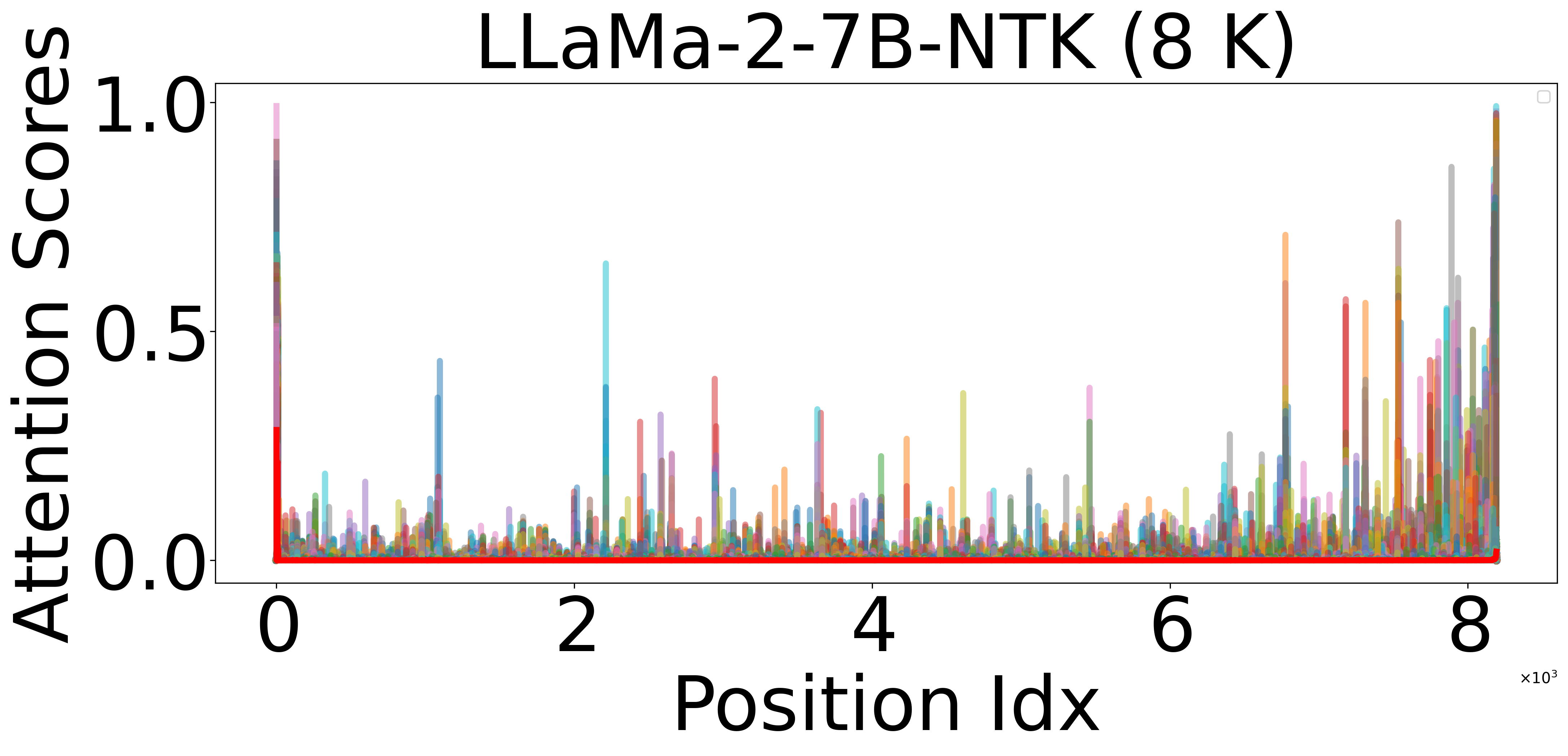}\label{fig_ntk_7b_8k_attn}}
    \vspace{-3mm}
    \caption{Attention distributions of RoPE, PI, YaRN, and NTK methods on 2K and 8K sequences. The red line represents the mean attention scores across all heads, layers, and examples. The other lines indicate the attention scores for each head in each layer.
    }
    \label{fig_ppl_attn_dist}
\end{figure*}

\begin{figure*}[!htbp]
    \centering
    \subfigure[RoPE on 2K sequences.]{\includegraphics[width=0.315\textwidth]{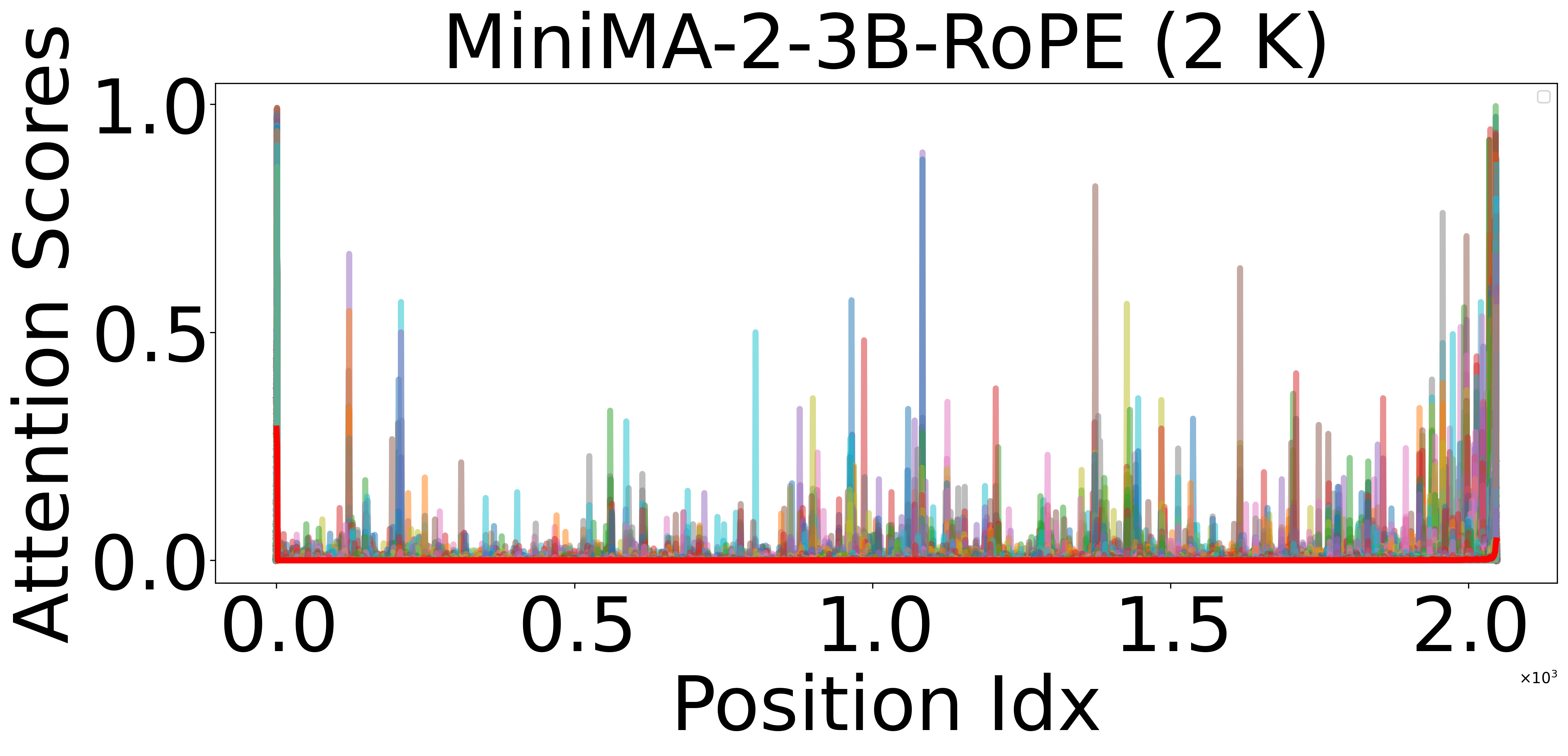}}
    \subfigure[RoPE on 8K sequences.]{\includegraphics[width=0.315\textwidth]{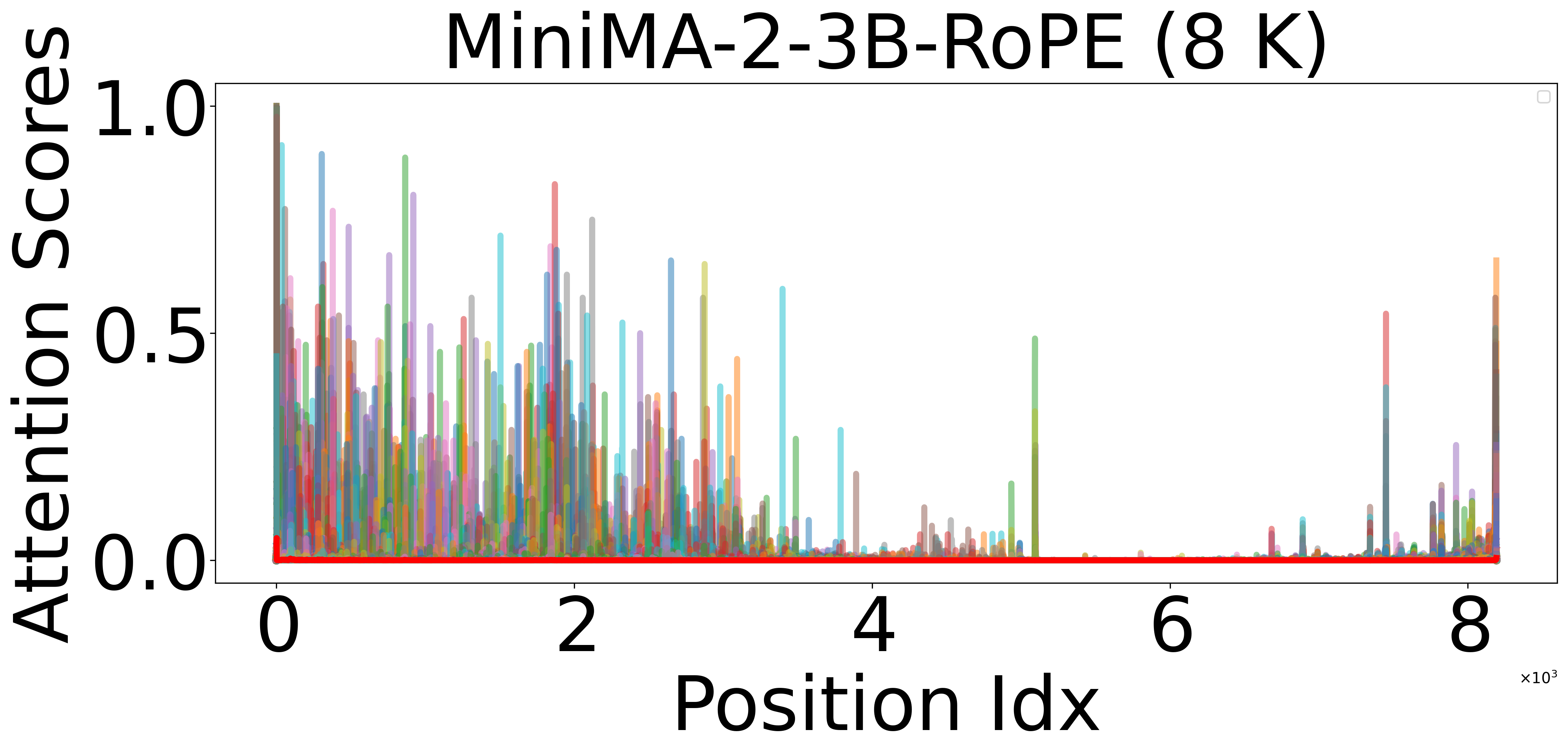}}
    \subfigure[PI on 8K sequences.]{\includegraphics[width=0.315\textwidth]{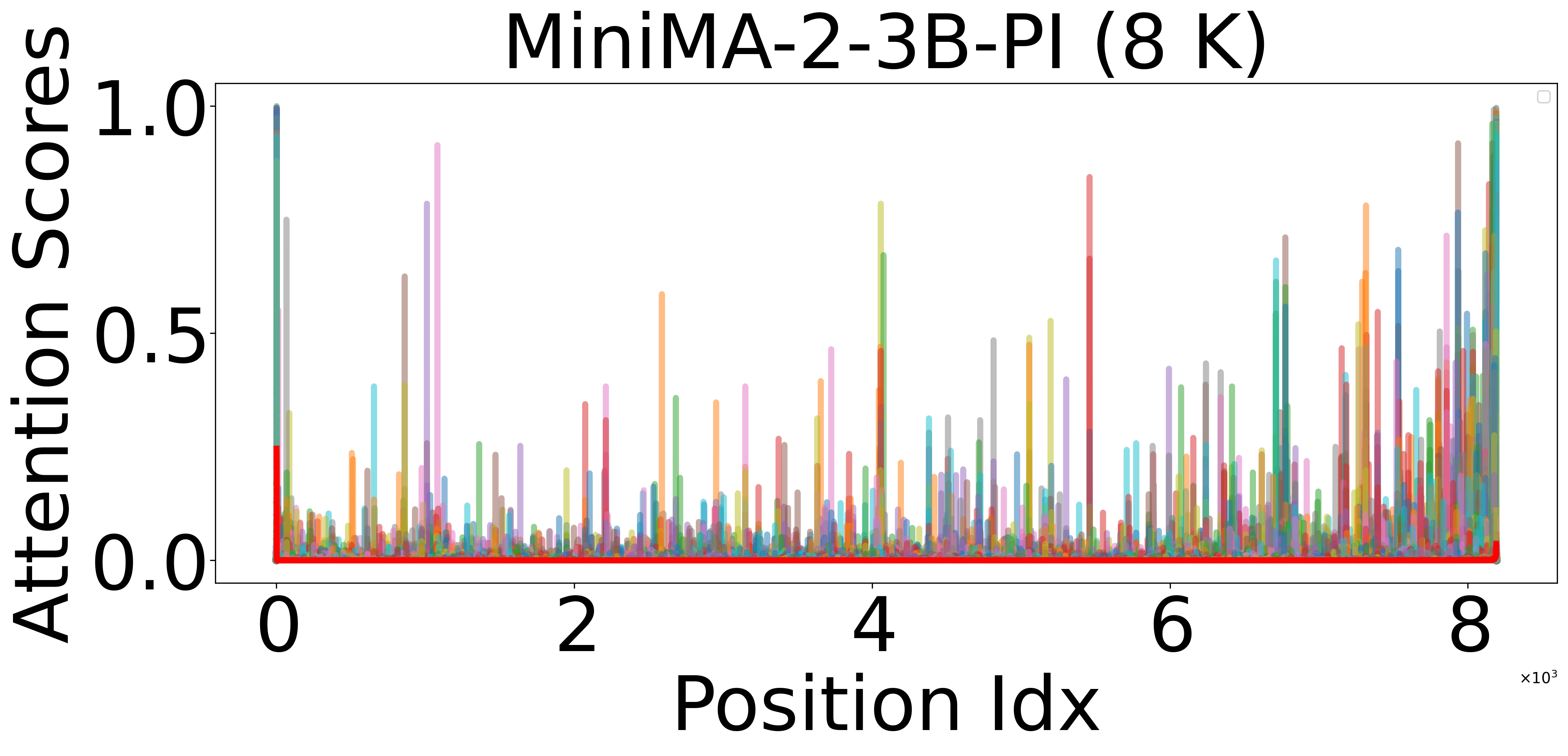}}
    
    \subfigure[YaRN on 8K sequences.]{\includegraphics[width=0.325\textwidth]{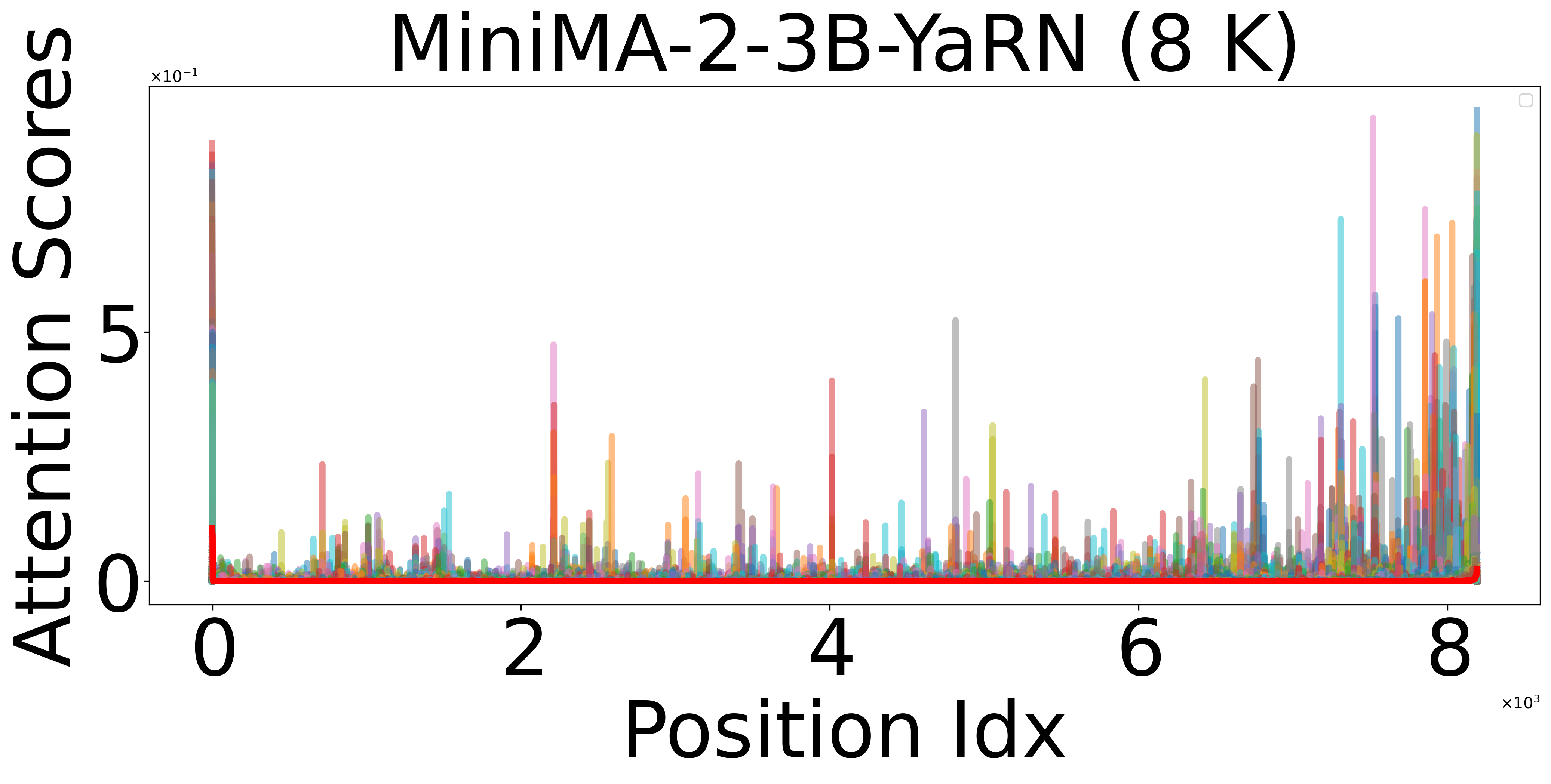}
    }
    \subfigure[NTK on 8K sequences.]{\includegraphics[width=0.34\textwidth]{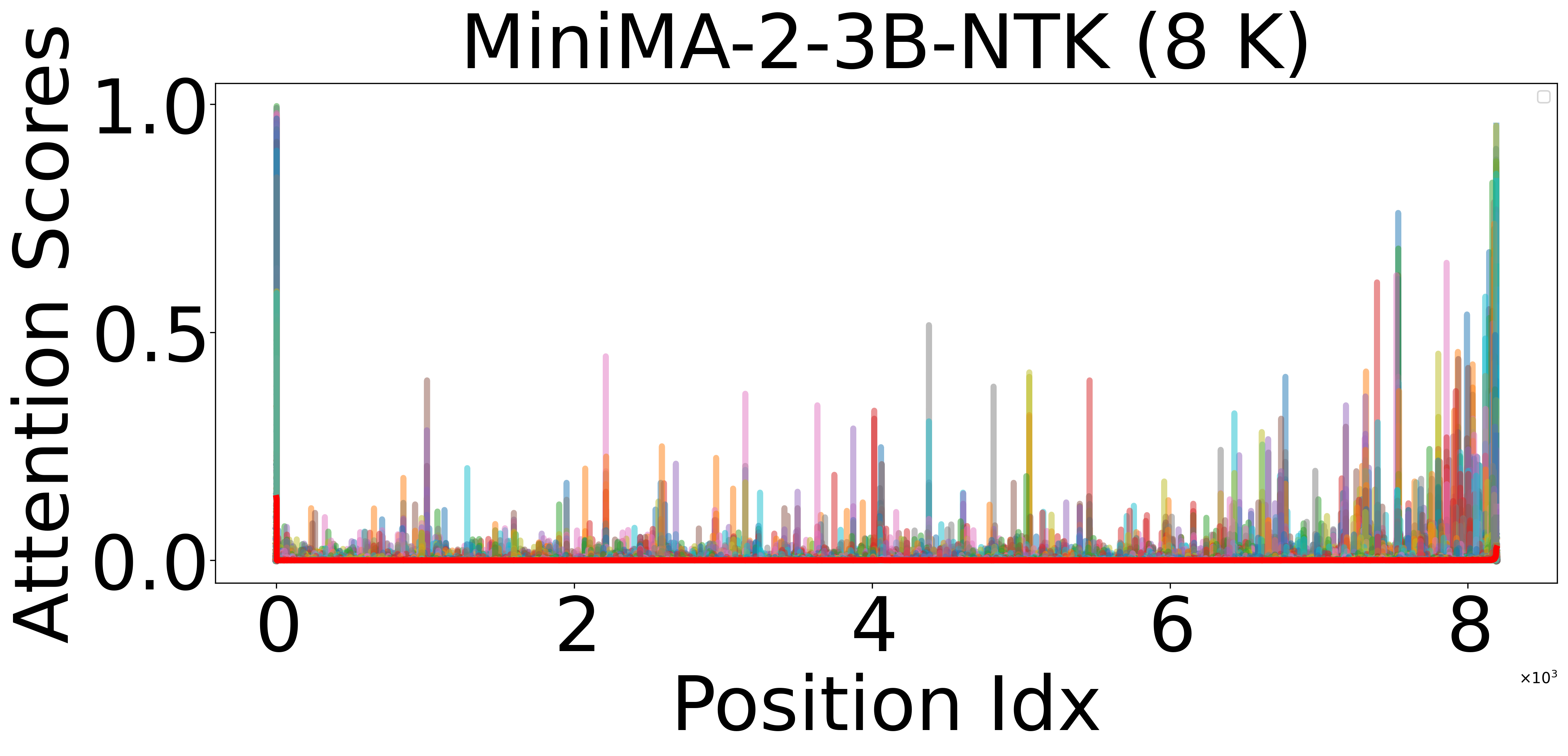}}
    \vspace{-3mm}
    \caption{Attention distributions of RoPE, PI, YaRN, and NTK methods on 2K and 8K sequences on MiniMA-2-3B.}
    \label{attn_dist_minima}
\end{figure*}

We study RoPE extensions by comparing performance on long-context perplexity testing. From the test, as illustrated in Figure \ref{fig_val_ppl}, we identify that NTK can extrapolate from 4K to 128K, whereas PI and YaRN can extrapolate to 62K. 
We observe similar results in both the smaller model MiniMA-2-3B and the larger model LLaMa-2-13B, as illustrated in Figures \ref{fig_val_ppl_minima} and \ref{fig_val_ppl_llama2_13b}.
To recognize why these RoPE extensions enable train-short-and-test-long properties in PPL, we collect the attention scores on 10 sequences in 2K and 8K and visualize their attention distributions. 
The followings are a few key takeaways from the attention perspective:



\textbf{RoPE extensions maintain the original attention patterns.}
As shown in Figure \ref{fig_ppl_attn_dist}, similar to the findings from \citet{chen2023extending}, we observe that the attention patterns fluctuate when the RoPE is tested on 8K sequences (exceeding the training length). However, with RoPE extensions, the attention distributions, as illustrated in Figures~\ref{fig_ppl_attn_dist}(c-e), revert to the original pattern seen in Figure \ref{fig_7b_2k_attn} when tested on 8K sequences.
Similar observations are seen in both LLaMa-2-13B and MiniMA-2-3B, as illustrated in Figures \ref{attn_dist_minima} and \ref{fig_ppl_attn_dist_llama13b}.

\begin{table}[t]
\centering
\resizebox{\linewidth}{!}{
\begin{tabular}{c|cccc}
\toprule
& PI & YaRN & NTK  & RoPE     \\ 
\midrule
LLaMa-2   & 1.29  & 0.05  & 0.06  & 0.00  \\ 
LLaMa-3  &	2.29 &	1.72  &	1.68 & 2.57 \\  
\bottomrule
\end{tabular}}
\caption{Jensen–Shannon (JS) divergence of mean attention distributions between different models at lengths of 2048 (top row) and 8192 (bottom row). A lower JS divergence indicates that the two attention distributions are similar.}
\label{tab_js_div}
\end{table}

\begin{figure*}[t]
    \centering
    \subfigure[RoPE]{\includegraphics[width=0.45\textwidth]{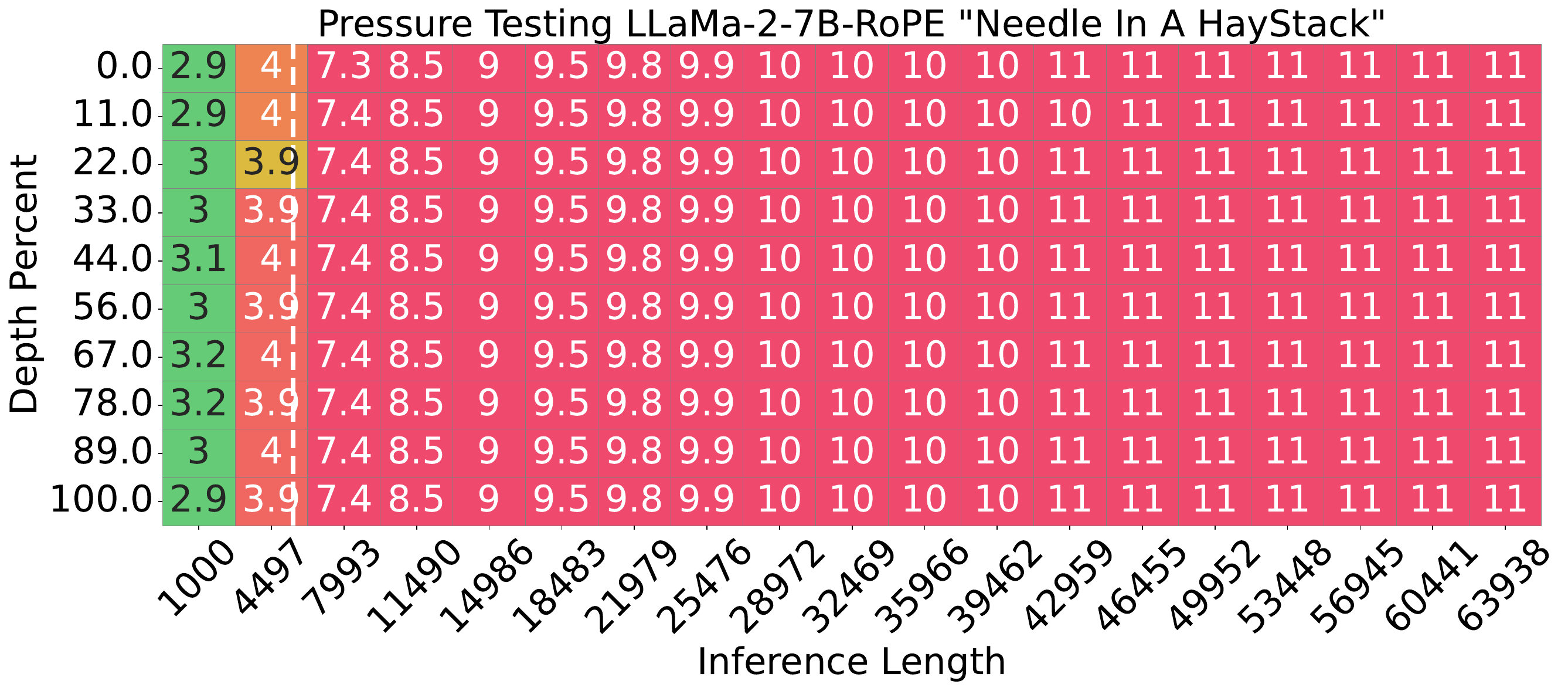}\label{fig_needle_baseline}}
    \subfigure[Finetuning with PI]{\includegraphics[width=0.45\textwidth]{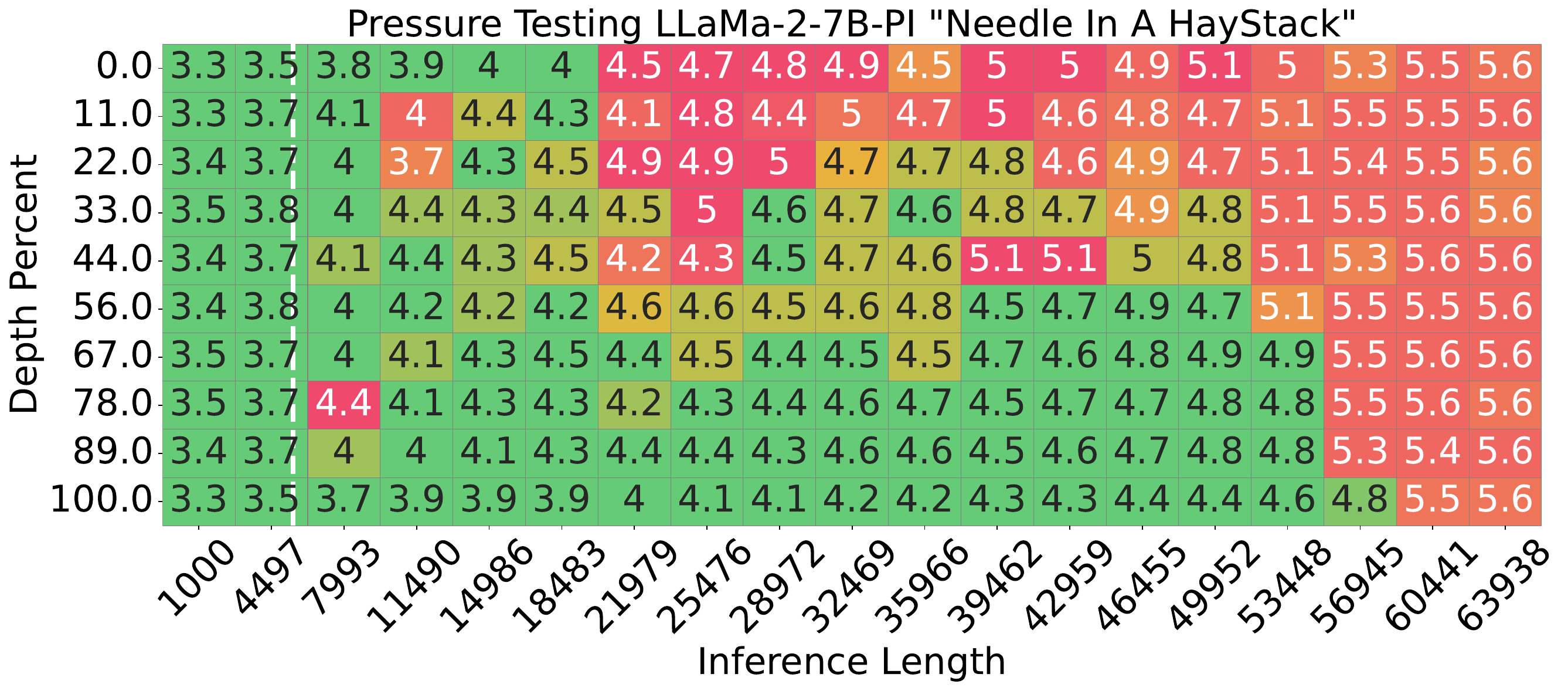}\label{fig_needle_linear_4k}
    }
    
    \subfigure[Finetuning with YaRN]{\includegraphics[width=0.45\textwidth]{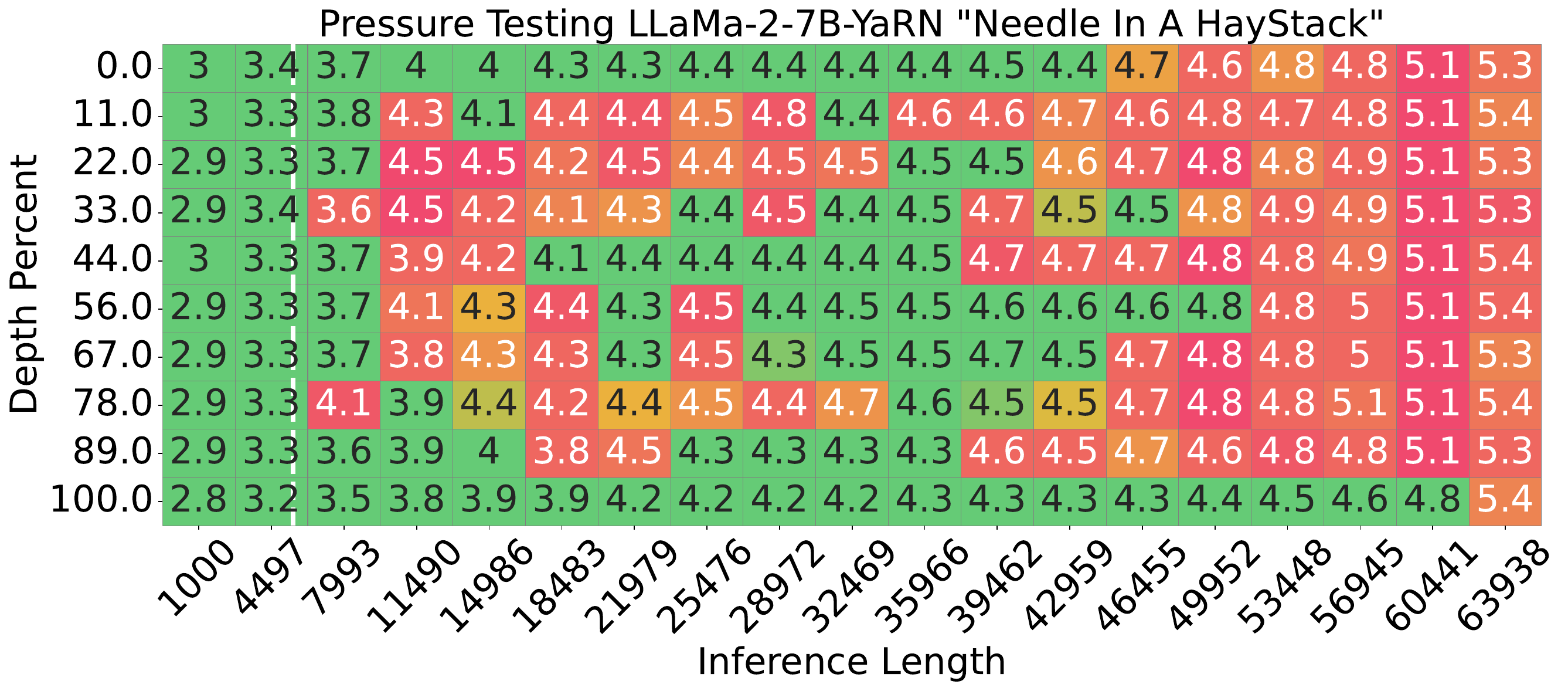}\label{fig_needle_yarn_4k}
    }
    %
    \subfigure[Finetuning(FT) with NTK]{\includegraphics[width=0.45\textwidth]{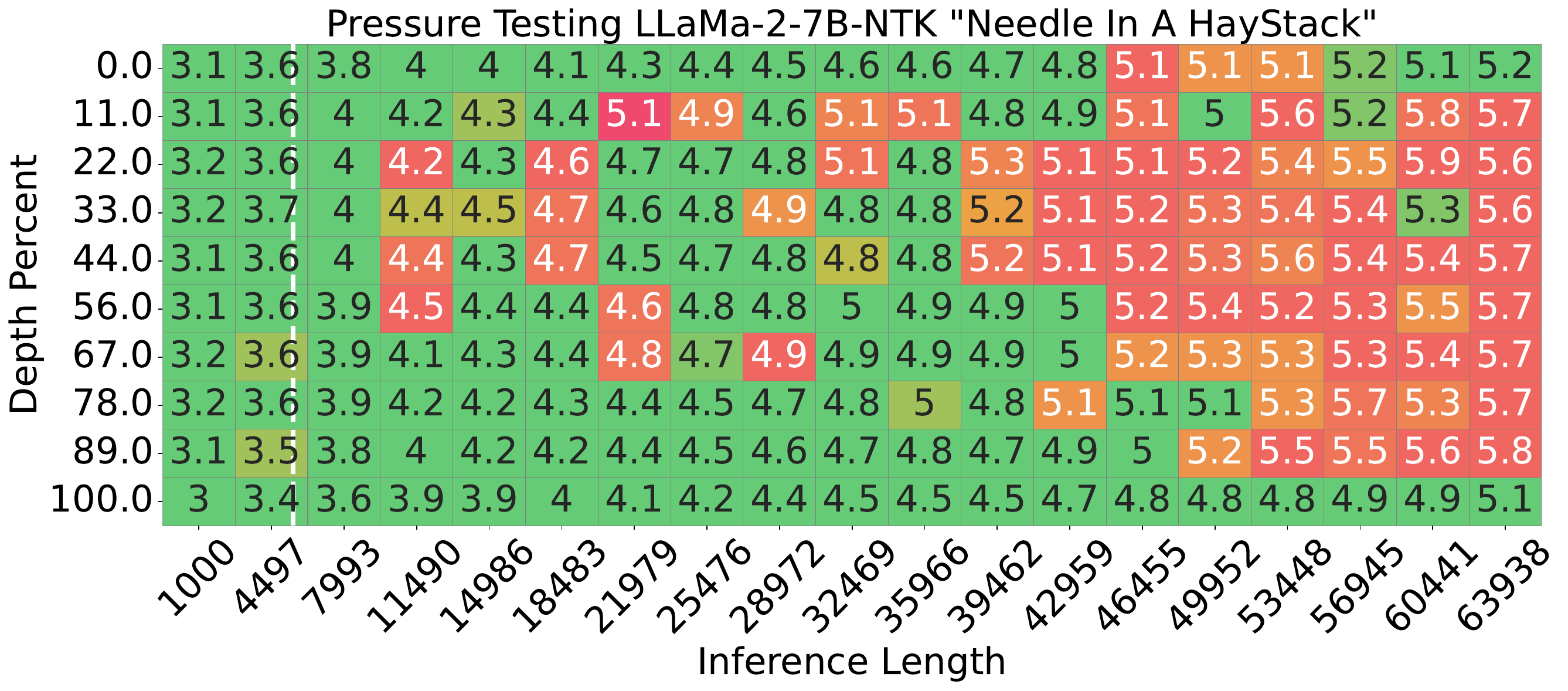}\label{fig_needle_ntk_4k}
    }
    
    \subfigure[FT on 4K with NTK from (d) ]{\includegraphics[width=0.45\textwidth]{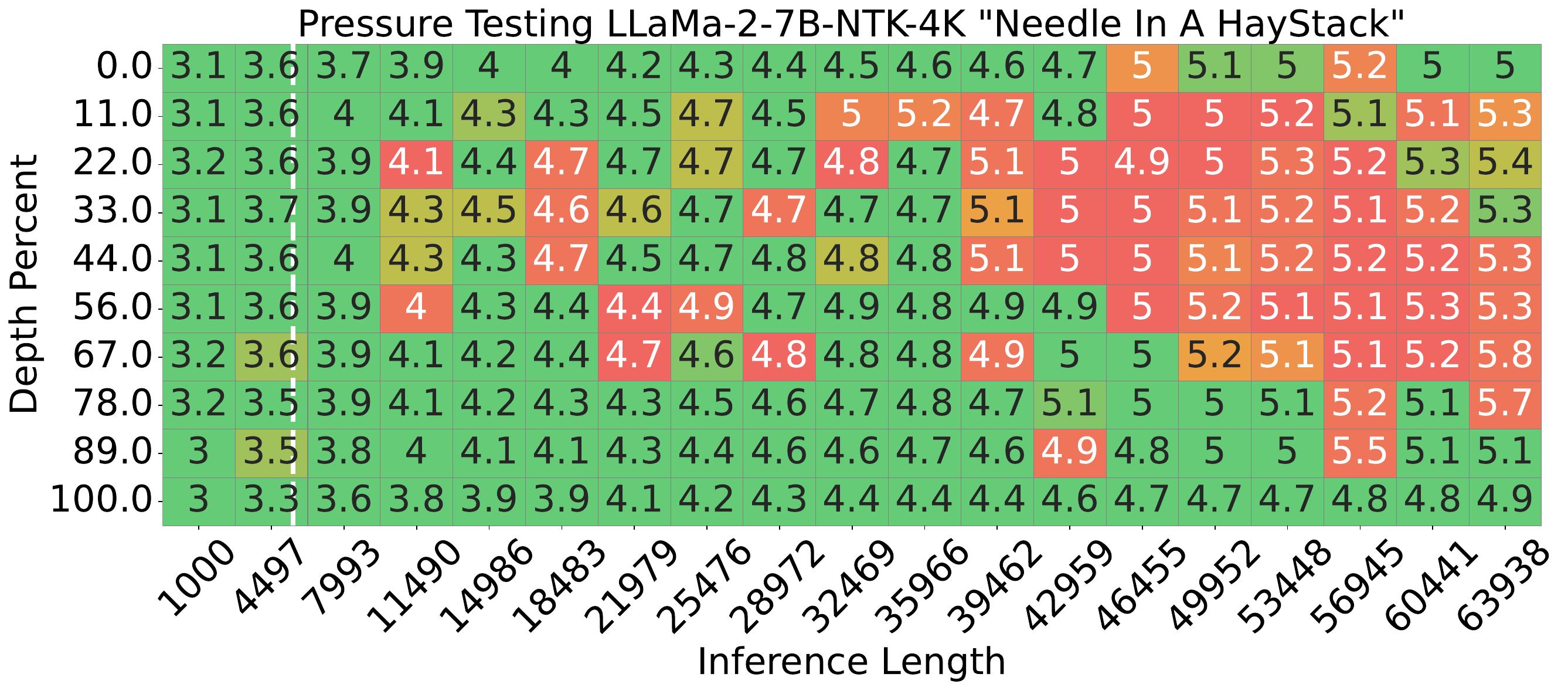}\label{fig_needle_ntk_4k_4k}}
    \subfigure[FT on 32K with NTK from (d)]{\includegraphics[width=0.45\textwidth]{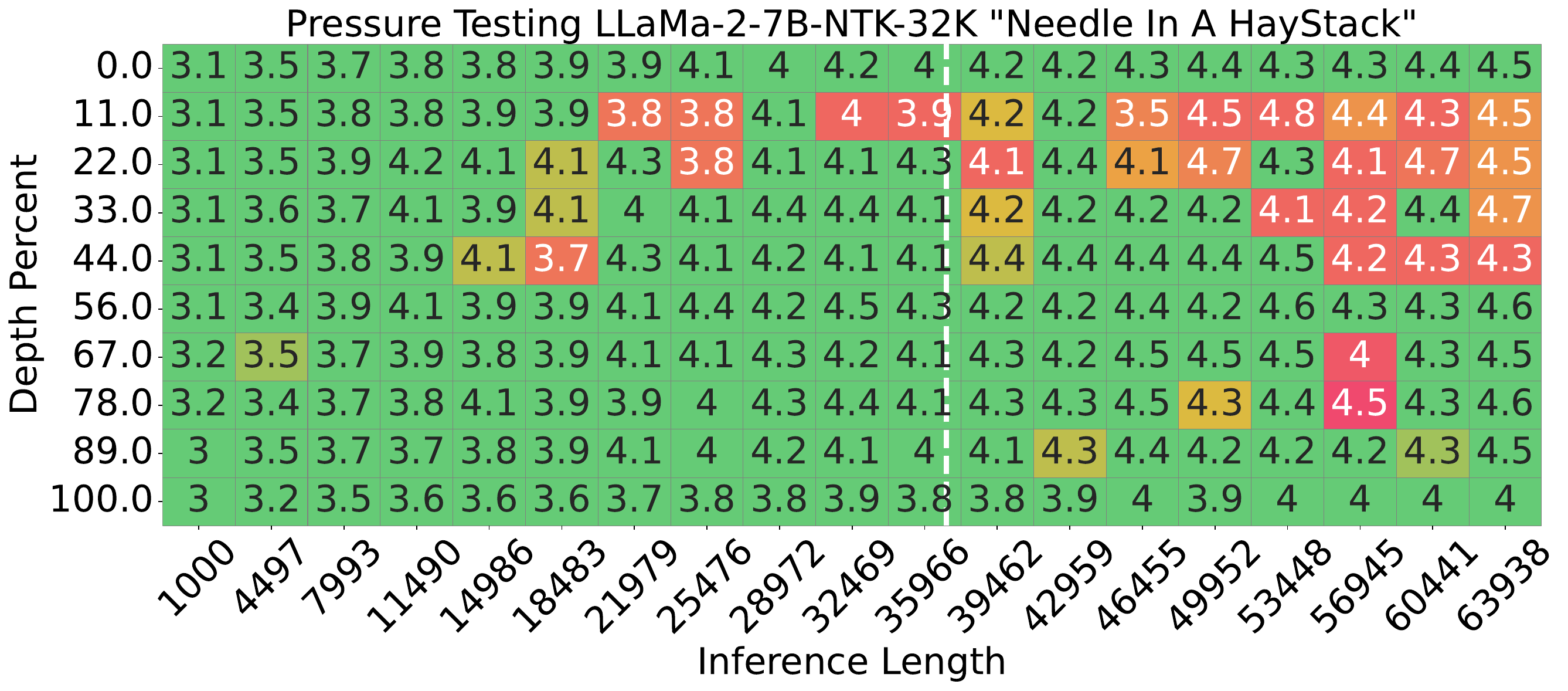}\label{fig_needle_ntk_4k_8k}
    }
    \vspace{-3mm}
    \caption{Performance comparison for the Needle-in-a-Haystack Test. The x-axis represents the length of the document, while y-axis indicates the depth percentage, showing the needle's position within the document.
    For instance, a position of 50\%  signifies that the needle is placed in the middle of the document. 
    A red cell indicates that the model fails to recall the information in the needle, whereas a green cell indicates success. 
    A white dashed line denotes the model's continual pretrain length. 
    Each value in the cells signifies the mean attention entropy, with higher values reflecting more dispersed attention.}
    \label{fig_needle}
\end{figure*}

\textbf{RoPE extensions closely resemble the attention patterns of models trained on longer context.}
To further verify whether RoPE extensions maintain the original attention patterns, we aim to directly quantify the Jensen–Shannon (JS) divergence between different attention distributions. 
Using LLaMa-2 and LLaMa-3 as baselines, we collected 10,240 samples of attention distributions to calculate the JS divergence.
As illustrated in the bottom row of Table \ref{tab_js_div}, the JS divergence between the RoPE extensions and LLaMa-3 is more minor than between the RoPE and LLaMa-3. 
This indicates that the attention patterns of RoPE extensions resemble those of models directly trained on a longer context.



\textbf{NTK and YaRN do not affect the attention patterns within the pretrained length.}
Some RoPE extensions can degrade performance within the original pretrained length \citep{peng2023yarn,zhang2024extending}. 
To verify whether RoPE extensions alter the attention patterns within the pretrained length, we also calculate the JS divergence among these models' attention distributions at a 2K length. 
As illustrated on the top row of Table \ref{tab_js_div}, the JS divergence for the NTK and YaRN is very low, almost zero, indicating minimal impact on attention distribution. On the contrary, the JS divergence for the PI is significantly higher. 
Therefore, we conclude that the NTK and YaRN methods do not affect attention patterns within the pretrained length.

\section{RoPE Extensions on Needle}
\label{rope_on_needle}
To understand the performance and behavior of the RoPE extensions on more challenging long-context tasks, we conduct Needle testing \cite{fu2024data}. 
As shown in Figure~\ref{fig_needle}(a-d), LLaMa-2-7B with RoPE extensions can pass more needle tests than the RoPE. 
However, as the context length increases, some tests fail, resulting in needle retrieval errors.
Eventually, almost all fail in extremely long contexts.
We also conduct Needle testing on the MiniMA-2-3B and LLaMa-2-13B models with RoPE and PI. Unlike the LLaMa-2-7B, the PI method shows a more significant improvement in the LLaMa-2-13B, 
as depicted in Figure \ref{fig_needle_llama13b}. In contrast, on the MiniMA-2-3B, PI passes only a few needle tests at longer lengths, as illustrated in Figure \ref{fig_needle_minima}. We attribute these observations to the impact of model size.
Below are key takeaways from the attention perspective:

\textbf{Attention uncertainty leads to more needle retrieval errors.}
To find the reason behind the needle retrieval errors, we calculate the entropy of attention for each length and depth, as illustrated in Figures \ref{fig_needle}. 
For details on the calculation of attention entropy, please refer to Appendix~\ref{detal_attnetion_entropy}.
Our findings demonstrate that the locations of needle retrieval errors often coincide with high attention entropy. For example, at the same depth, the positions with errors are among the top-k in entropy; similarly, at the same length, the error positions also have high entropy.
We hypothesize that the increase in attention entropy with longer test lengths is due to the train-short-and-test-long setting. During inference, the number of tokens handled by the self-attention mechanism far exceeds that during training. More tokens lead to more dispersed attention, i.e., higher uncertainty, causing a mismatch between training and inference.

\textbf{A natural approach to lower attention uncertainty for enhancing extrapolation.}
A direct solution is to train on longer contexts, thereby increasing the number of attention tokens during training and reducing attention uncertainty. 
To validate our hypothesis, we finetune models on 4K and 32K training lengths with the same tokens on NTK.
As shown in Figures \ref{fig_needle_ntk_4k_4k} and \ref{fig_needle_ntk_4k_8k}, compared to models trained in short contexts, models trained in more extended contexts exhibited significantly lower attention uncertainty. For example, at length 63938, the attention entropy is generally below 5. 
The Needle test pass rates improved significantly, especially in longer testing contexts. Conversely, models trained with the same number of tokens but shorter context sizes showed little to no change in attention entropy, remaining similar to the original one (\ref{fig_needle_ntk_4k}).


\section{Conclusions}
This paper provides the first thorough understanding of RoPE extensions for long-context LLMs from an attention perspective, evaluated on two widely-used benchmarks: Perplexity and Needle-in-a-Haystack.
Extensive experiments demonstrate some valuable findings: 
1) Compared to direct extrapolation, RoPE extensions can maintain the original training length attention patterns.
2) Large attention uncertainty leads to retrieval errors in needle testing in RoPE extensions.
3) Using longer continual pretraining lengths for RoPE extensions can reduce attention uncertainty and significantly enhance extrapolation in target LLMs.

\section*{Limitations}
This paper primarily analyzes the widely-used decoder-only LM, LLaMa~\citep{touvron2023llama}. It does not include a validation study of encoder-decoder and encoder-only architectures.

\section*{Acknowledgements}
We would like to thank the anonymous reviewers and meta-reviewer for their insightful suggestions.
This work was supported by the National Natural Science Foundation of China under Grant U23B2055 and 62276077, and Shenzhen Science and Technology Program under Grant ZDSYS20230626091203008.

\bibliography{custom}

\appendix

\section{Experimental Results on MiniMA-2-3B}
\label{minima_2_3b}


    

\begin{figure}[H]
    \centering
    \subfigure[RoPE]{\includegraphics[width=0.45\textwidth]{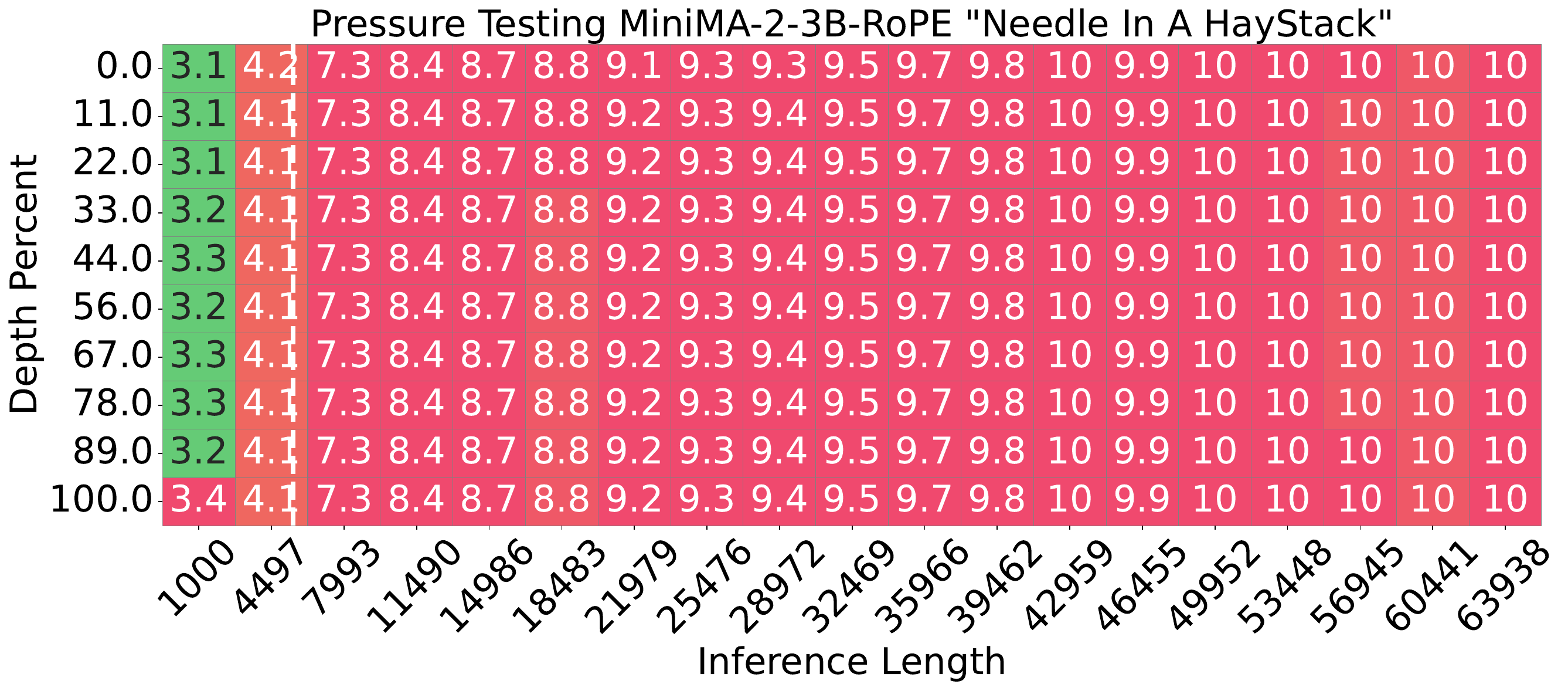}}
    
    \subfigure[Finetuning with PI]
    {\includegraphics[width=0.45\textwidth]{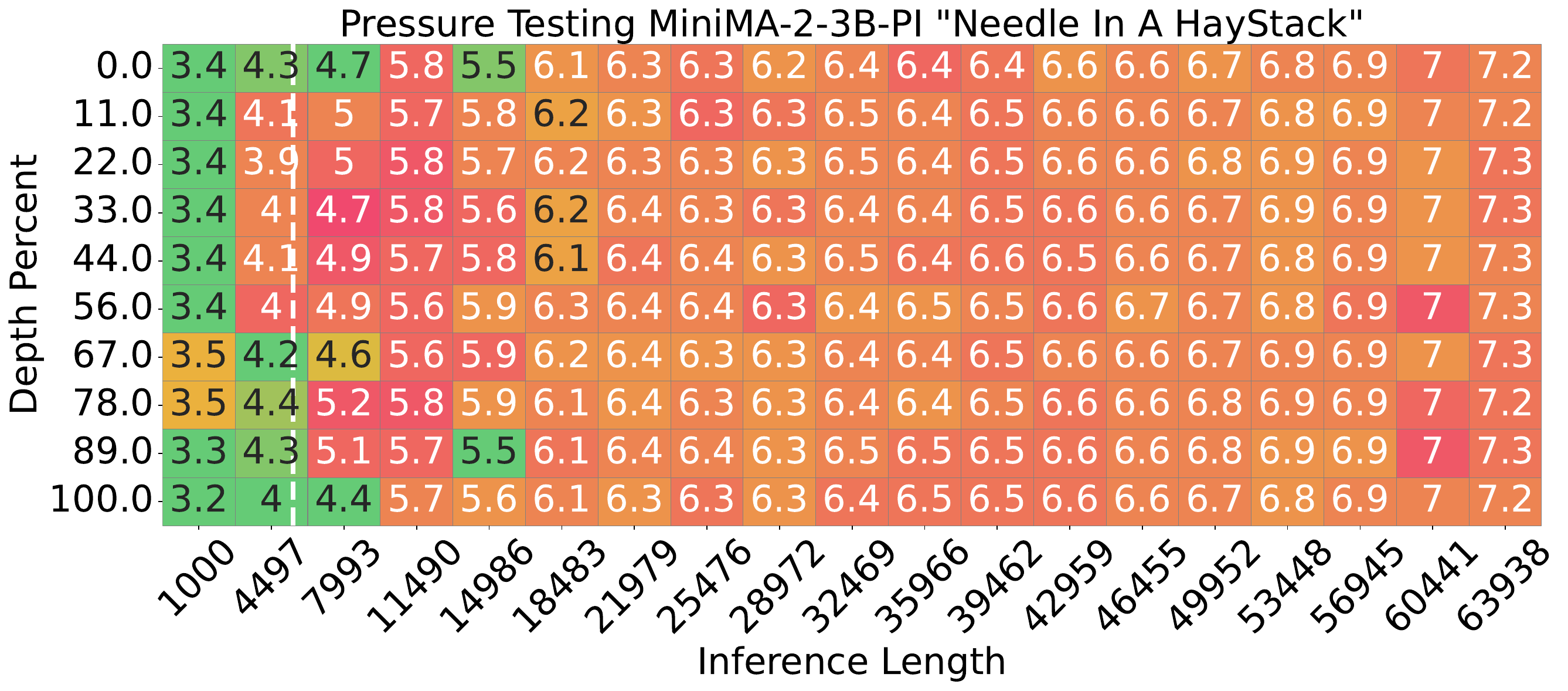}
    }
    \vspace{-3mm}
    \caption{Performance comparison for the Needle-in-a-Haystack Test of MiniMa-2-3B.}
    \label{fig_needle_minima}
\end{figure}

Similar to the analysis in \S~\ref{rope_on_needle}, the Needle-in-a-Haystack Test for MiniMa-2-3B also indicates that the locations of needle retrieval errors frequently align with areas of high attention entropy.

\section{Experimental Results on LLaMa-2-13B}
\label{llama_2_13b}


\begin{figure*}[!htbp]
    \centering
    \subfigure[RoPE on 2K sequences.]{\includegraphics[width=0.315\textwidth]{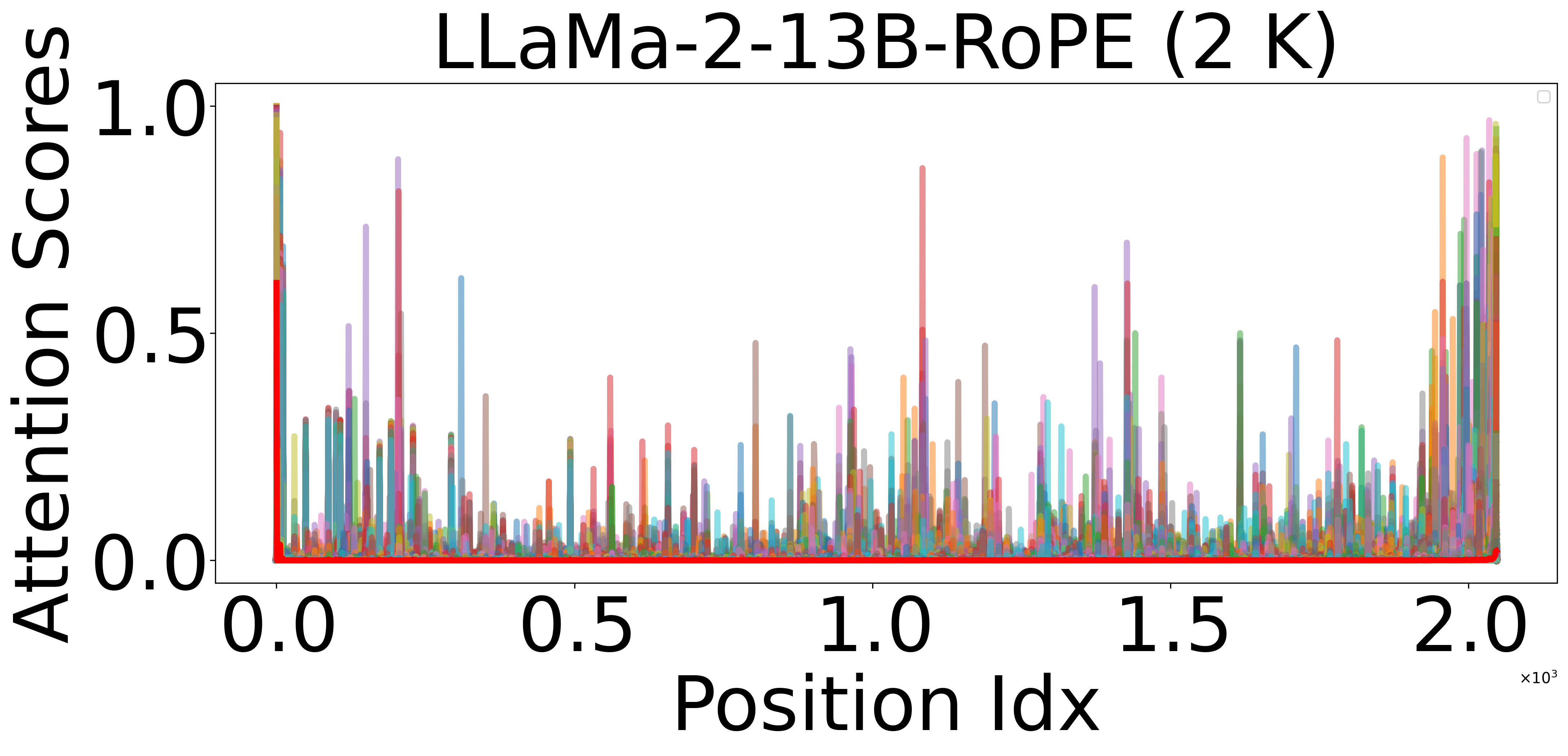}}
    \subfigure[RoPE on 8K sequences.]{\includegraphics[width=0.315\textwidth]{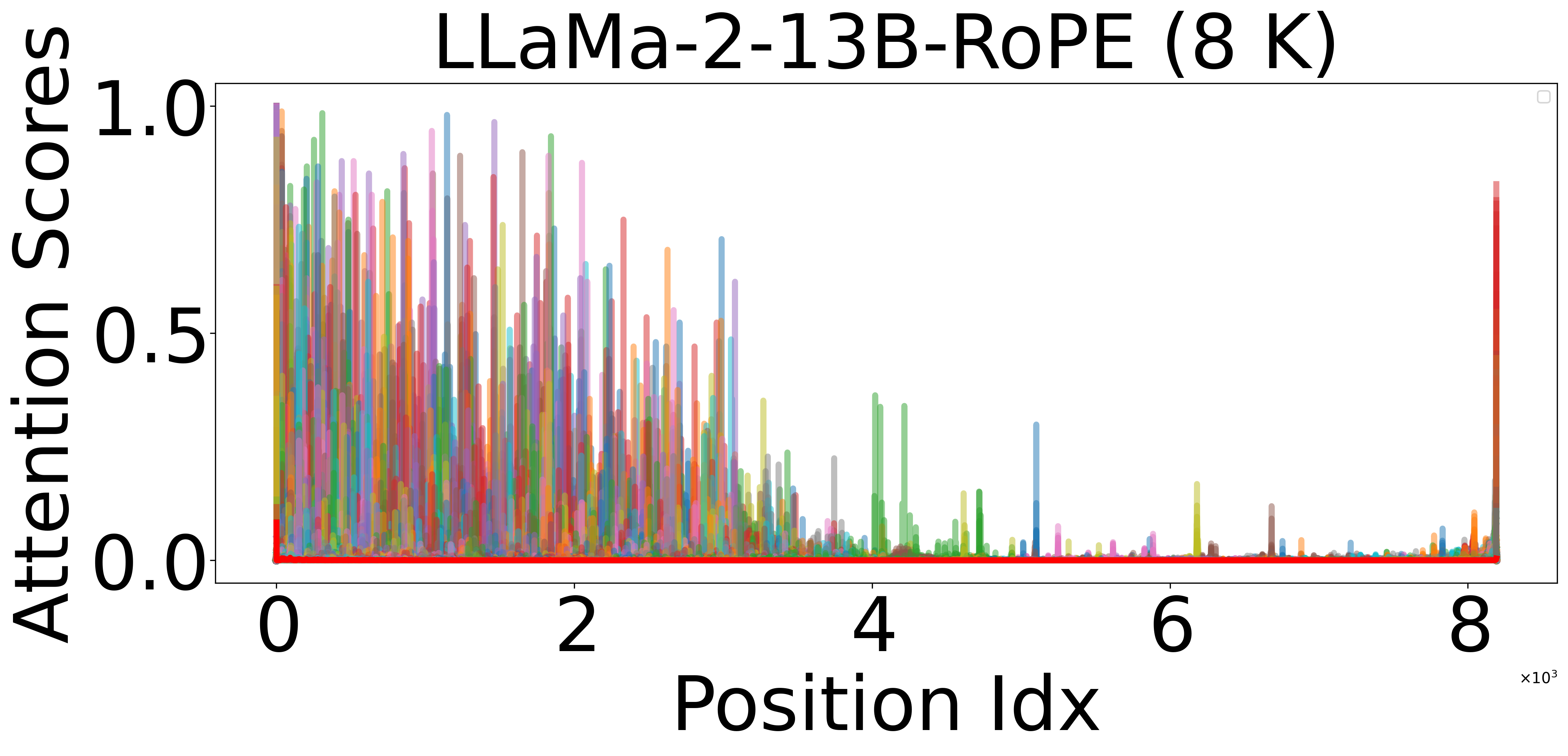}}
    \subfigure[PI on 8K sequences.]{\includegraphics[width=0.315\textwidth]{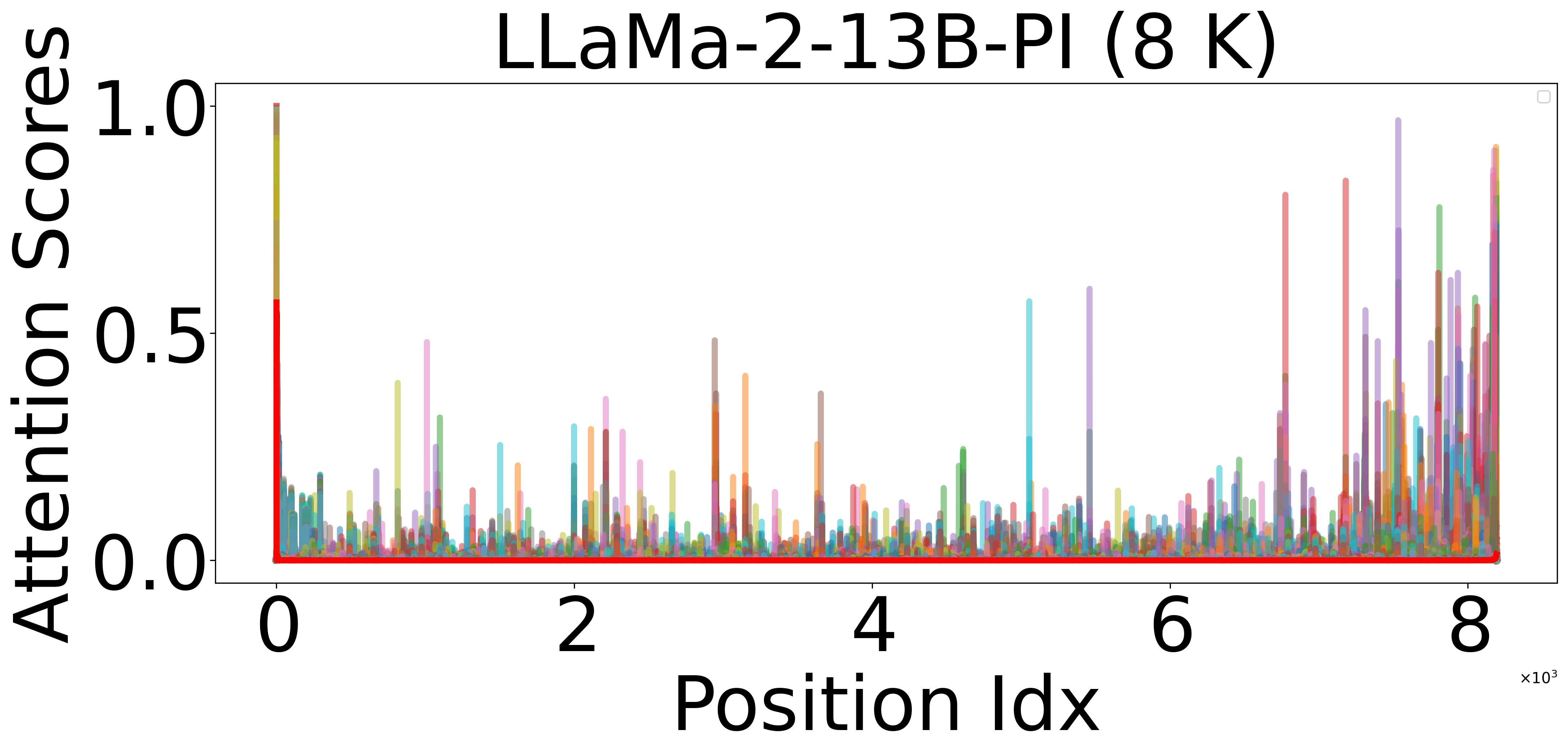}}

    \subfigure[YaRN on 8K sequences.]{\includegraphics[width=0.315\textwidth]{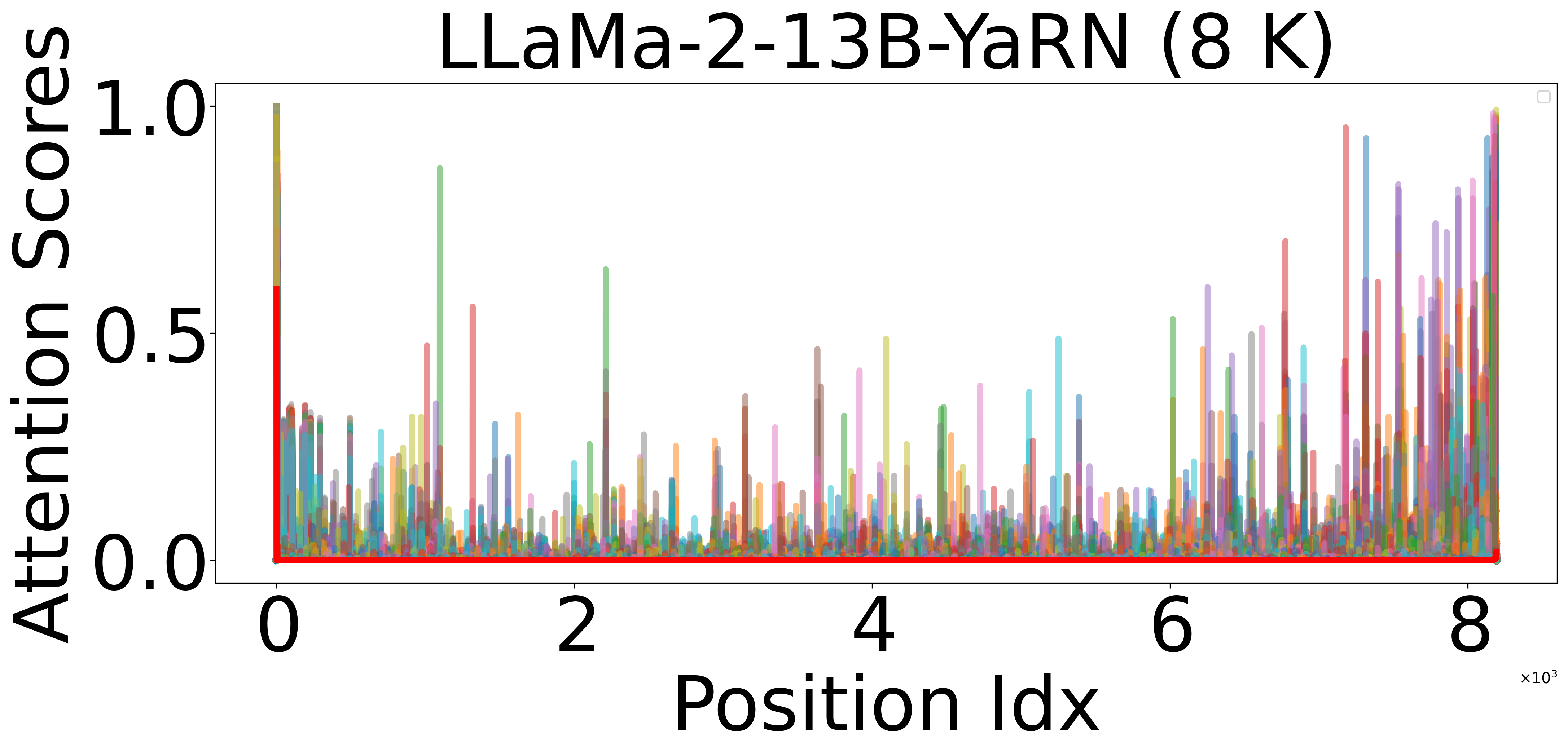}
    }
    \subfigure[NTK on 8K sequences.]{\includegraphics[width=0.315\textwidth]{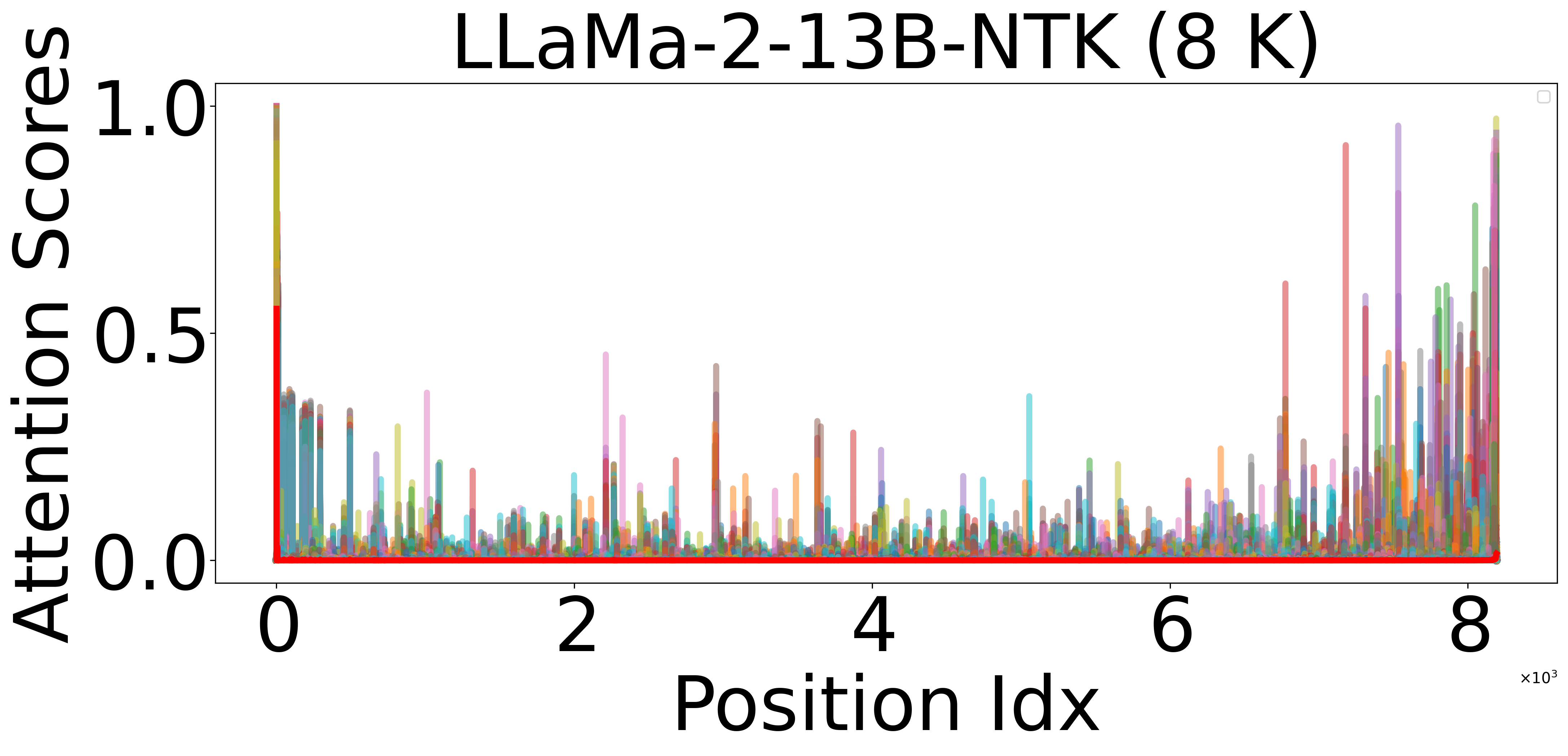}}
    \vspace{-3mm}
    
    \caption{Attention distributions of RoPE, PI, YaRN, and NTK methods on 2K and 8K sequences on LLaMa-2-13B.}
    \label{fig_ppl_attn_dist_llama13b}
\end{figure*}
Consistent with the analysis in \S~\ref{rope_on_ppl}, we observe that the attention patterns fluctuate when RoPE is applied to 8K sequences, which exceed the training length. However, when using RoPE extensions, the attention distributions return to their original patterns for 8K sequences, as demonstrated in Figures~\ref{fig_ppl_attn_dist_llama13b}.

\begin{figure}[H]
    \centering
    \subfigure[RoPE]{\includegraphics[width=0.45\textwidth]{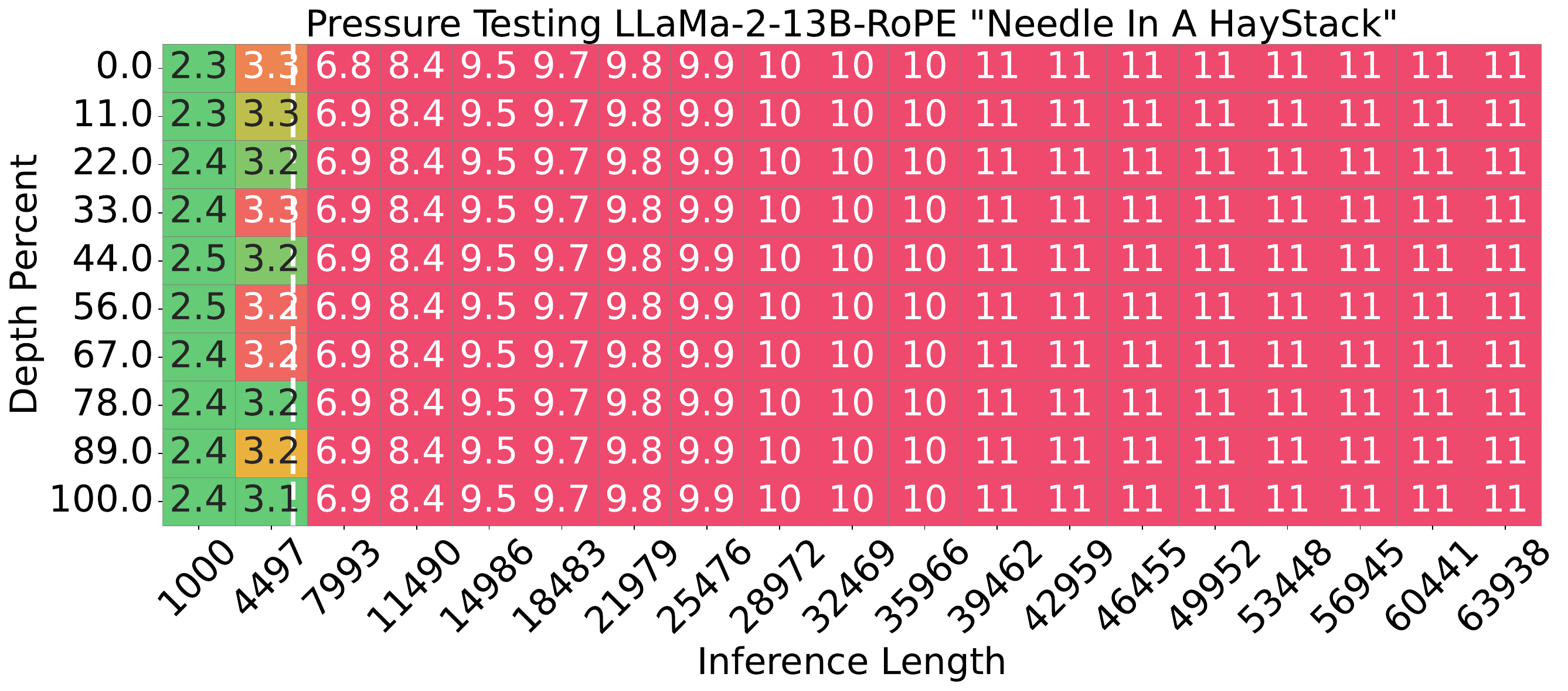}}
    
    \subfigure[Finetuning with PI]
    {\includegraphics[width=0.45\textwidth]{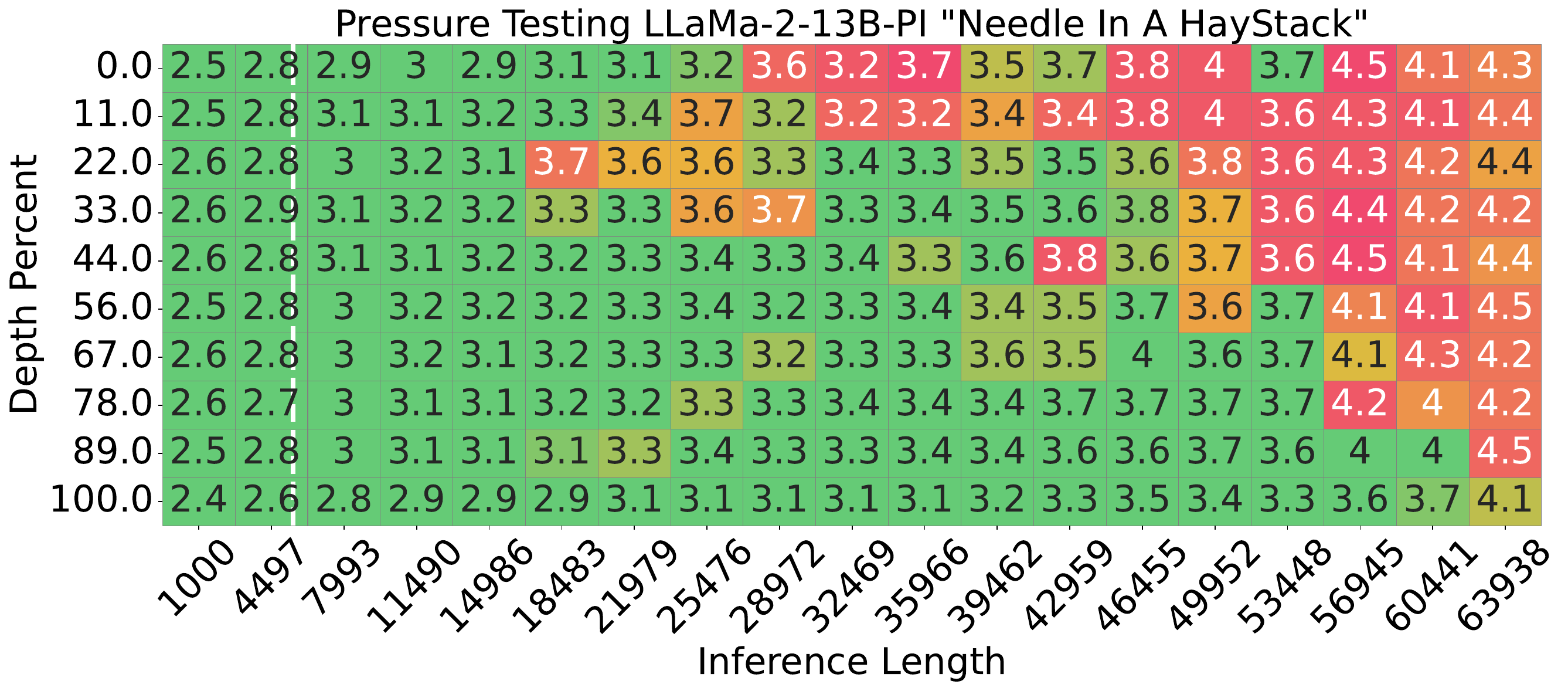}
    }
    \vspace{-3mm}
    \caption{Performance comparison for the Needle-in-a-Haystack Test of LLaMa-2-13B.}
    \label{fig_needle_llama13b}
\end{figure}
Similar to the analysis in \S~\ref{rope_on_needle}, the Needle-in-a-Haystack Test for LLaMa-2-13B also indicates that the locations of needle retrieval errors frequently align with areas of high attention entropy.

\section{Detailed Calculation of Attention Entropy}
\label{detal_attnetion_entropy}









\begin{algorithm*}[!htbp]
\caption{Calculation of Attention Entropy}
\label{alg:attention_entropy}
\begin{algorithmic}[1]

\State \textbf{Input:} model, prompt
\State \textbf{Output:} average attention entropy score

\Procedure{AttentionEntropy}{$model, prompt$}
\State Initialize $entropy\_list \gets []$
\State $output\_tokens \gets \text{[  ]}$ 
\While{not end of generation}
  \State $token, attention\_distribution \gets \text{GenerateTokenAndGetAttention}(model, prompt \,+\, output\_tokens)$
  \State $output\_tokens \text{.append}(token)$
  \State $entropy \gets \text{CalculateEntropy}(attention\_distribution)$
  \State $entropy\_list \text{.append}(entropy)$
\EndWhile

\State $average\_entropy \gets \text{Average}(entropy\_list)$
\State \Return $average\_entropy$
\EndProcedure

\Function{CalculateEntropy}{$distribution$}
\State $entropy \gets 0$
\ForAll {$p \text{ in } distribution$}
  \If {$p > 0$}
    \State $entropy \gets entropy - p \log(p)$
  \EndIf
\EndFor
\State \Return $entropy$
\EndFunction

\end{algorithmic}
\end{algorithm*}

\end{document}